\documentclass[12pt]{article}
\usepackage[utf8]{inputenc}

\usepackage{amsmath}
\usepackage{amssymb}
\usepackage{amsthm,lineno}
\usepackage{graphicx}
\usepackage{esint}
\usepackage{float}
\usepackage{indentfirst}
\usepackage{authblk}
\usepackage{amsfonts}
\usepackage{mathrsfs}
\usepackage{graphicx}
\usepackage{booktabs}
\usepackage{subcaption}
\usepackage{bm}
\usepackage{bbm}
\usepackage{wrapfig, enumitem}
\usepackage[colorlinks = true,citecolor=blue]{hyperref}
\usepackage{color}
\usepackage{mathtools}
\usepackage[round, sort]{natbib}
\usepackage{algorithm}
\usepackage{algorithmic}
\usepackage[a4paper,margin=1in]{geometry}
\usepackage[toc,page]{appendix}
\usepackage{comment}
\usepackage{diagbox}
\usepackage{makecell}
\usepackage{threeparttable}
\usepackage{multirow}
\usepackage{titlesec}
\usepackage{tikz}
 
\usepackage{tocloft}

\titlespacing*{\section}{0pt}{3pt plus 3pt minus 1pt}{3pt plus 3pt minus 1pt}
\titlespacing*{\subsection}{0pt}{3pt plus 3pt  minus 1pt}{3pt plus 3pt minus 1pt}

\newtheorem{definition}{Definition}

\newcommand{\bs}{\boldsymbol}

\newcommand{\R}{\mathbb{R}}
\newcommand{\N}{\mathcal{N}}
\newcommand{\E}{\mathbb{E}}

\makeatletter
\newcommand*\bigcdot{\mathpalette\bigcdot@{.6}}
\newcommand*\bigcdot@[2]{\mathbin{\vcenter{\hbox{\scalebox{#2}{$\m@th#1\bullet$}}}}}
\makeatother

\title{Differentially Private Normalizing Flows for Density Estimation, Data Synthesis, and Variational Inference with Application to Electronic Health Records}
\author[]{Bingyue Su\footnote{co-first authors}}
\author[]{Yu Wang$^*$}
\author[]{Daniele E. Schiavazzi}
\author[]{Fang Liu}
\affil[]{Department of Applied and Computational Mathematics and Statistics, University of Notre Dame, Notre Dame, IN}

\date{}

\begin{document}
\maketitle\vspace{-24pt}
\begin{abstract}
\noindent Electronic health records (EHR) often contain sensitive medical information about individual patients, posing significant limitations to sharing or releasing EHR data for downstream learning and inferential tasks. We use normalizing flows (NF), a family of deep generative models, to estimate the probability density of a dataset with differential privacy (DP) guarantees, from which privacy-preserving synthetic data are generated. We apply the technique to an EHR dataset containing patients with pulmonary hypertension. We assess the learning and inferential utility of the synthetic data by comparing the  accuracy in the prediction of the hypertension status and variational posterior distribution of the parameters of a physics-based model.  In addition, we use a simulated dataset from a nonlinear model to compare the results from variational inference (VI) based on privacy-preserving synthetic data, and privacy-preserving VI obtained from directly privatizing NFs for VI with DP guarantees given the original non-private dataset. 
The results suggest that synthetic data generated through differentially private density estimation with NF can yield good utility at a reasonable privacy cost. We also show that VI obtained from differentially private NF based on the free energy bound loss may produce variational approximations with significantly altered correlation structure, and loss formulations based on alternative dissimilarity metrics between two distributions might provide improved results.

\vspace{6pt}

\noindent \textbf{keyword}: differential privacy, privacy-preserving, synthetic data, normalizing flow, variational inference, free energy bound, KL divergence, density estimation, electronic health records 
\end{abstract}


\section{Introduction}

\subsection{Background}\label{sec:background}
In the last decade, data science has gained tremendous popularity as an interdisciplinary field, thanks to the exponential growth in data volume, fast development of powerful and efficient data analysis techniques, and availability of computational facilities and resources. When sharing data among researchers or releasing information to the public, there are always privacy concerns and risks for re-identification or disclosure of sensitive information. Even with key identifiers removed, an adversary may still be able to identify an individual or learn his or her sensitive information, leveraging publicly available data and sophisticated attacks. Some notable privacy attacks includes re-identification based on health records \citep{sweeney97, sweeney2013matching}, Netflix prize~\citep{netflix2}, membership attacks \citep{genotype, backes2016membership, shokri2017membership}, among others.

In this work, we use electronic health records (EHR) as an example to present approaches for privacy-preserving data sharing and analysis. EHR data often contain sensitive medical information about individual patients. For example, cardiovascular EHRs, the type of EHR data examined in this work, contain measurements of vital signs such as heart rate, arterial and venous, systemic and pulmonary pressures,  cardiac output, atrial and ventricular pressures and volumes, ejection fraction, etc., which can be used to diagnose hypertension. 
Due to privacy concerns, accessing large single-center EHR datasets for research purposes is often problematic~\citep[see recent efforts in making large clinical datasets publicly available, for example, the MIMIC-IV dataset discussed in][]{johnson2020mimic}, not to mention merging multiple EHR data sources to create large multi-center datasets enabling complex predictive models to be trained~\citep[see, in this context, the eICU Collaborative Research Database discussed in][]{pollard2018eicu}. 

To mitigate privacy concerns when releasing information and sharing data, various privacy concepts have been proposed over the past two decades,  such as $k$-anonymity~\citep{samarati1998protecting}, $t$-closeness~\citep{li2007t}, $l$-diversity~\citep{aggarwal2008general}, and differential privacy (DP)~\citep{dwork2006calibrating, dwork2006our, dwork2006delta}, among others. We focus on DP, a popular privacy concept and framework in contemporary privacy research. Besides providing mathematical guarantees on privacy and being robust to various privacy attacks ~\citep{dwork2017exposed}, DP has attractive properties, such as privacy loss composition and immunity to post-processing, which have significantly contributed to its popularity for both research and practical applications.
Privacy-preserving techniques and algorithms are available for a wide spectrum of statistical and machine-learning methods with DP guarantees, ranging from releasing counts and histograms to predictions from regression models or deep neural networks.

In this work, we examine differentially private normalizing flow (NF)  to generate synthetic data via density estimation and to obtain variational inference (VI) with DP guarantees, respectively, under a pre-defined privacy budget. NFs are generative models  \citep{rezende2015variational, dinh2016density, kingma2016improved, papamakarios2017masked} with major applications  in density estimation and VI. 
%

\subsection{Our Work and Contributions}

Our work comprises two parts.  The first part focuses on  synthetic data generation from differentially private density estimation and the utility of the synthetic data for downstream leaning tasks.
To our knowledge, there is only one other paper~\citep{waites2021differentially} on incorporating DP in NF for density estimation. In addition, no prior work has explored the use of synthetic data generated through DP-NF for downstream learning such as prediction or inference.  Our work intends to fill this gap. 

Specifically, we apply DP-NF for density estimation given an EHR dataset that contains hemodynamic measurements from a cohort of patients including individuals suffering from elevated pulmonary pressures~\citep[group II pulmonary hypertension, see][]{simonneau2013updated} and heart failure with preserved ejection fraction~\citep{harrod2021ehr}.
The EHR dataset has a small sample size and a relatively large number of attributes with continuous or discrete numerical measurements and contains missing values. 
This provides an opportunity to examine the feasibility of DP-NF in generating useful privacy-preserving synthetic data for small-sized datasets, under data type heterogeneity and a greater degree of realism compared to simulated data or well-studied benchmark datasets. We run regression and classification on synthetic data generated by DP-NF and compare their prediction accuracy with that obtained from the original data. 
We also perform VI of a computationally expensive physics-based model given the synthetic data. To reduce the computation cost, we leverage an offline-trained and fast neural network surrogate model for the physics-based model. 

The second part of this work investigates privacy-preserving VI by directly privatizing NF for VI with DP guarantees (as opposed to VI from private synthetic data from differentially private density estimated via NF).
We examine the performance of the procedure in inferring the parameters of a nonlinear regression model in simulated data. 
To the best of our knowledge, this paper is the first to examine the feasibility of differentially privatizing NF for VI using free energy bound loss and stochastic gradient-based optimization.
We show that this loss formulation may yield differentially private variational distributions that deviate significantly from the target density, particularly on the correlations if the the target distribution is multivariate, when the marginal variances in the differentially private variational distributions are inflated due to DP guarantees compared to the target. We also examined several other dissimilarity metrics as the loss function for NF-VI, similar observations as in the case of free energy bound were observed in some of them, but not all of them. This phenomenon suggests the importance of carefully selecting the loss function when performing VI based on DP-NF.

In both parts, we apply Gaussian DP to achieve privacy guarantees through stochastic gradient descent optimization, in contrast to the moment account method~\citep{abadi2016deep} for tracking privacy loss used in existing work~\citep{waites2021differentially}. 
Gaussian DP yields a tighter bound for the privacy loss composition and is shown to yield higher prediction accuracy deep learning~\citep[see][]{bu2020deep}, compared to the moment account technique that is based on R\'enyi DP.

\subsection{Related Work}\label{sec:work}
A straightforward approach for differentially private density estimation from which synthetic data can be generated is through histogram sanitization with DP, such as the perturbed histogram or the smoothed histogram approaches \citep{wasserman2010statistical}. Histograms as density estimators suffer from the curse of dimensionality, a problem that is only exacerbated after factoring in DP; in addition, continuous attributes need to be discretized in histogram estimators.  \citet{hall2013differential} and  \citet{alda2017bernstein} develop DP mechanisms to privatize functional outputs with applications to kernel density estimation. Adaptations of local DP to non-parametric density estimation have also been proposed~\citep{butucea2020local, kroll2021density}. 
\citet{waites2021differentially} developed the privacy-preserving version of NF for density estimation by privatizing SGD iteratively with the moment accountant method~\citep{abadi2016deep} to track privacy loss. 

Work also exists on VI with privacy guarantees. \citet{karwa2015private} directly model a DP mechanism used for sanitizing a likelihood function and obtain a variational approximation to the resulting sanitized posterior distribution.
\citet{jalko2016differentially} privatize SGDs using the moment accountant method with an assumed family of variational distributions. 
\citet{park2016variational} develop a privacy-preserving variational Bayes framework by sanitizing the expected sufficient statistics over the distribution of latent variables using the Gaussian mechanism when Bayesian models belong to the conjugate-exponential family.
\citet{sharma2019differentially} examine differentially private VI in the setting of federated learning. 
\citet{jalko2022dpvim} propose aligned gradients to factor in the fact that different elements in the gradient vector associated with multi-dimensional variational parameters are often of different magnitudes, aiming to improve the utility of differentially private VI without incurring additional privacy cost. To our knowledge, none of the work above realizes differentially private VI through NF.

Data synthesis for statistical disclosure limitation was first proposed by~\citet{rubin1993statistical} and is employed by government agencies for releasing survey data and by the healthcare industry to share medical data, among other applications. Synthesized surrogate data sets have the same structure as the original dataset but comprise pseudo-individuals. Differentially private data synthesis (DIPS) provides a solution to integrate formal privacy guarantees into data synthesis. DIPS can be achieved through both model-free and model-based approaches; interested readers can refer to~\citet{bowen2020comparative} for an overview of some of the recent DIPS techniques. 
The introduction of variational autoencoders~\citep[VAE][]{kingma2013auto}, generative adversarial networks~\citep[GANs][]{goodfellow2014generative}, and NF~\citep{rezende2015variational,papamakarios2017masked}, opened new possibilities for data synthesis through deep-neural-networks-based generative models with DP guarantees. 
\citet{xie2018differentially} generate privacy-preserving synthetic data via GANs (DP-GANs), incorporating DP guarantees using the moment accountant method. \citet{jordon2018pate} combine GANs and the Private Aggregation of Teacher Ensembles (PATE) framework~\citep{papernot2016semi, papernot2018scalable} to generate differentially private synthetic data and demonstrate its superiority over DP-GANs in utility in empirical studies.
\citet{chen2018differentially} propose a differentially private autoencoder-based generative model and a differentially private variational autoencoder-based generative model and demonstrate both approaches can yield satisfactory utility but the former is more robust against some privacy attacks than the latter.
\citet{beaulieu2019privacy} apply DP-GANs to release synthetic data to a real clinical dataset with reasonable DP guarantees and demonstrate the utility of the synthetic data. 
\citet{pfitzner2022dpd} generate privacy-preserving synthetic data by aggregating only the decoder component of VAEs in federated learning with DP while leaving the encoders personal at local servers.

The rest of the paper is organized as follows. Section \ref{sec:prelim} introduces the definitions of NF for density estimation and VI and the concept of DP. Section \ref{sec:DP-NF-DE} presents the procedure for differentially private NF for density estimation and generation of privacy-preserving synthesis from estimated density, with an application to a real EHR dataset. Section \ref{sec:DP-NF-VI} presents the procedure for differentially private NF for VI, with an application to repeated measures data simulated from a nonlinear model and an in-depth analysis of the utility of variational distributions obtained from the procedure. The paper concludes in Section \ref{sec:discussion} with a summary and some final remarks.

\section{Preliminaries}\label{sec:prelim}

\subsection{Normalizing Flow}\label{sec:nf}

Normalizing Flow (NF) is defined as a map $F: \R^d \times \Lambda\to \R^d$ parameterized by $\lambda \in \Lambda$ transforming realizations from an easy-to-sample base distribution such as $\bm z_0 \sim \N(\bm{0}, I_d)$, to realizations from a desired target density. 
Specifically, $F$ consists of a composition of $K$ bijections $F_k: \R^d \times \Lambda_k\to \R^d$, each parameterized by $\lambda_k \in \Lambda_k$: $F_{\lambda}(\bm z_0) = F(\bm z_0; \lambda) = \left[F_K( \ \cdot \ ; \lambda_K)\circ F_{K-1}( \ \cdot \ ; \lambda_{K-1})\circ \cdots \circ F_1( \ \cdot  \ ;\lambda_1)\right](\bm z_0)$, where $\bm z_k = F_k(\bm z_{k-1}; \lambda_k)$ for $k=1,\ldots,K$. Since $F_k( \ \cdot \ ;\lambda_k)$ is a bijection from $\bm z_{k-1}$ to $\bm z_{k}$, $q_k(\bm z_k)$, the distribution of $\bm z_k$, can be obtained by the change of variable
\begin{equation}\label{eqn:changeOfVar}
q_k(\bm z_k) = q_{k-1}(\bm z_{k-1})\left|\det \frac{\partial F_k^{-1}}{\partial \bm z_{k-1}}\right| = q_{k-1}(\bm z_{k-1})\left|\det \frac{\partial F_k}{\partial \bm z_{k-1}}\right|^{-1}.
\end{equation}
Taking the logarithm and summing over $k$, Eqn.~\eqref{eqn:changeOfVar} becomes
\begin{equation}\label{eqn:loglikelihood}
\textstyle\log q_K(\bm z_K) = \log q_0(\bm z_0) - \sum_{k=1}^K \log \left|\det \frac{\partial F_k}{\partial \bm z_{k-1}}\right|.
\end{equation}

Generally, the objective of training a NF is to determine an \emph{optimal} set of parameters $\lambda$ so that $q_{K}$ can approximate either a target density $p$ (VI) or the distribution of the observed data $\bs x$ (density estimation). 

For general bijections, the computational cost of computing the Jacobian determinants in Eqn.~\eqref{eqn:changeOfVar} may increase significantly with the dimensionality $d$ and the number of layers $K$. To efficiently compute these determinants, various NF formulations have been proposed based on coupling layers such as RealNVP~\citep{dinh2016density} and GLOW~\citep{kingma2018glow}, and autoregressive transformations such as MAF~\citep{papamakarios2017masked} and IAF~\citep{kingma2016improved}, both of which are associated with tractable triangular Jacobian matrices and easy to compute.

In the applications of NF to density estimation and VI in Sections~\ref{sec:DP-NF-DE} and \ref{sec:DP-NF-VI}, we use MAF~\citep{papamakarios2017masked}. The autoregressive property of MAF is obtained by setting $p(z_i|z_{1},\dots,z_{i-1}) = \phi((z_i - \mu_i) / e^{\alpha_i}$, where $\phi$ is the density function of the standard normal distribution, $\mu_i = f_{\mu_i}(z_{1},\dots,z_{i-1})$, $\alpha_i = f_{\alpha_i}(z_{1},\dots,z_{i-1})$, and $f_{\mu_i}$ and $f_{\alpha_i}$ are masked autoencoder neural networks~\citep[MADE,][]{germain2015made}.
Consider $\bm z$ and $\hat{\bm z}$ as the input and output of a MADE network having $L$ hidden layers with $d_l,\,l=1,\cdots, L$ neurons per layer. 
The mappings between layers could be represented as
\[
\left\{ \begin{matrix*}[l]
\bm{y}_{1}\!\!\!\!& \!=\! h_1(b_1 + (\bm W^1 \odot \bm M^1)\bm z), & \text{where } \bm M^1 \text{ is }d_1\times d \text{ and }\bm M^1_{u, v} = \mathbbm{1}_{m^{1}(u) \geq v}\\[0.7em]
\bm{y}_{l}\!\!\!\!& \!=\! h_l(b_l + (\bm W^l \odot \bm M^l)\bm{y}_{l-1}), & \text{where } \bm M^l \text{ is }d_{l}\times d_{l-1} \text{ and }\bm M^l_{u, v}\!=\!\mathbbm{1}_{m^{l}(u)\geq m^{l-1}(v)}\\[0.7em]
\hat{\bm{z}}\!\!\!\! &\!\!=\! h_{L+1}(b_{L+1} + \!(\bm W^{L+1} \odot \bm M^{L+1})\bm{y}_{L}) & \text{where } \bm M^{L+1} \text{ is }d \times d_{L} \text{ and }\bm M^{L+1}_{u, v} = \mathbbm{1}_{u > m^L(v)},\\
\end{matrix*}\right.
\]
where $h_l$ is the activation function between layer $l-1$ and $l$ for $l=1,\cdots, L+1$, $m^l(k)$ is a pre-set or random integers from $1$ to $d-1$, $b_l$ and $\bm W^l$ are the bias and weight parameters for layer $l=1,\dots,L+1$.

The matrix product $\prod_{l=L+1}^{1}\bm M^{l}$  is strictly lower triangular, satisfying the autoregressive property~\citep{germain2015made}. 
The masks $\bm M^{1},\dots,\bm M^{L+1}$ enable the computation of all $\mu_i$ and $\alpha_i$ in a single forward pass~\citep{papamakarios2017masked}. Additionally, the Jacobian of each MAF layer is lower-triangular, with determinant equal to $|\det \partial f / \partial \bm z |^{-1} = \exp(\sum_{i=1}^{d} \alpha_i)$. 
Before each MADE layer, batch normalization~\citep{ioffe2015batch} can be used to normalize the outputs from the previous layer. Batch normalization performs elementwise scaling and shifting operations, 
is easily invertible, has a tractable Jacobian, and is found to help stabilize and accelerate training and increase accuracy~\citep{papamakarios2017masked}.

When NF is employed for density estimation given observed data $\bs x$, the following log-likelihood is maximized, assuming $n$ independent observations $\bs{x}_{i},\,i=1,\dots,n$.
\begin{align}\label{eqn:MLE}
\notag\textstyle\ell(\lambda; \bm x) & =\textstyle \log q_K(\bm x) =\sum_{i=1}^n \log q_K(\bm x_i)\\ 
& = \sum_{i=1}^n \log q_0(\bm{z}_{i, 0}) - \sum_{i=1}^n\sum_{k=1}^K \left|\det \frac{\partial F_k}{\partial \bm z_{i,k-1}}\right|,
\end{align}
where $\bm z_{i, k-1} = F^{-1}_k \circ\cdots\circ F^{-1}_{K-1} \circ F^{-1}_K (\bm x_i)$.
Once the NF parameters are obtained via the maximum likelihood approach, samples from $q_K$ can be generated by first sampling from the base distribution and then transforming the samples through a sequence of bijections. NF has been shown to provide accurate approximations of complex density functions of various data types~\citep{papamakarios2017masked, dinh2016density, kingma2018glow}.

The other major application of NF is VI. 
Consider the likelihood  $l(\bs z; \bs x)$  of parameters or latent variables $\bs z$ given observed data $\bs x_i$ for $i=1,\ldots, n$. 
Given prior $p(\bs z)$, an NF-based approximation $q_K(\bs z)$ of the posterior distribution $p(\bs z|\bs x)$ can be computed by maximizing the lower bound to the log marginal likelihood $\log p(\bs x)$ (the \emph{evidence lower bound} or ELBO), or, equivalently, by minimizing the free energy bound~\citep{rezende2015variational}, expressed as
\begin{align}
\mathcal{F}(\bm x)& = \E_{q_K(\bm z)}\left[\log q_K(\bm z) - \log p(\bm x, \bm z)\right] = \E_{q_K(\bm z_K)}\left[\log q_K(\bm z_K) - \log p(\bm x, \bm z_K)\right]\notag\\
& = \E_{q_0(\bm z_0)}[\log q_0(\bm z_0)] - \E_{q_0(\bm z_0)}[\log p(\bm x, \bm z_K)] - \E_{q_0(\bm z_0)}\left[\sum_{k=1}^K \log \left|\det \frac{\partial F_k}{\partial \bm z_{k-1}}\right|\right]\notag\\
&= \sum_{i=1}^n\!\left(\!\frac{1}{n}\E_{q_0(\bm z_0)}[\log q_0(\bm z_0)] - \E_{q_0(\bm z_0)}[\log p(\bm x_i, \bm z_K)] - \!\frac{1}{n}\E_{q_0(\bm z_0)}\!\left[\sum_{k=1}^K \log \left|\det \frac{\partial F_k}{\partial \bm z_{k-1}}\right|\right]\right)\notag\\
& = \textstyle \sum_{i=1}^n l_i\label{eqn:KL}
\end{align}
The expectations in Eqn.~\eqref{eqn:KL} can be approximated by their corresponding Monte Carlo (MC) estimates using samples $\bm{z}_{i,0},\,i=1,\dots,n$ from the basic distribution $q_0$ during implementation. 

\subsection{Differential Privacy}

Differential Privacy (DP) is a mathematical framework designed to produce robust privacy guarantees. It aims at protecting privacy of every individual in a dataset in the worst-case scenario. 
\begin{definition}[$(\epsilon,\delta)$-DP]\citep{dwork2006calibrating,dwork2006our,dwork2006delta}\label{def:dp1}
A randomized mechanism $\mathcal{M}$ is $\epsilon$-DP, if $\Pr[\mathcal{M}\left(D_{1}\right)\in S]\leq e^{\epsilon}\Pr[ \mathcal{M}\left(D_{2}\right)\in S]+\delta$ for any $S\in {\rm range}(\mathcal{M})$ and neighbouring datasets $D_{1}, D_{2}$ that differ by one record, where $\epsilon>0$ and $\delta\in[0,1)$.
\end{definition}
Informally, DP implies that the chance of identifying an individual in a dataset using a randomized mechanism is limited if the released information is about the same with or without that individual in the data.
The quantities $\epsilon$ and $\delta$ represent privacy budget or privacy loss parameters.  
When $\delta=0$, $(\epsilon,\delta)$-DP in Definition~\ref{def:dp1} reduces to pure $\epsilon$-DP. The smaller the $\epsilon$, the higher the privacy protection there is for the individuals in the data.

When we apply DP mechanisms repeatedly to query a dataset, privacy loss will accumulate during the process. It is therefore necessary to track and account for the accumulated loss to so that the privacy loss would not be too large. 
This has motivated the development of variants and extensions of the original DP concepts, aiming at achieving tighter bounds for composite privacy loss, such as the concentrated DP \citep{cPD,bun2016concentrated, bun2018composable}, R\'{e}nyi DP \citep{mironov2017renyi}, and Gaussian DP \citep{dong2019gaussian}. 
\citet{abadi2016deep} propose a widely popular moment account technique based on R\'{e}nyi DP to track privacy loss in stochastic gradient descent-based optimization.  \citet{dong2019gaussian} provide a tighter bound based on Gaussian DP (GDP), which is the DP paradigm we will use for the reminder of this paper.

\begin{definition}[$\mu$-GDP]\citep{dong2019gaussian} Let $T$ be a trade-off function that maps $[0,1]$ to $[0,1]$, defined  as 
$T(P_{0},P_{1})(\alpha)=\inf\{\beta_{\phi}:\alpha_{\phi}\leq\alpha\} $
for two distributions $P_{1}$ from $P_{0}$ to be distinguished through hypothesis testing ($H_0:P_0$ vs. $H_1:P_1$), where $\phi\in[0,1]$ is a rejection rule with the type I error $\alpha_{\phi}=\mathbb{E}_{P_{0}}[\phi]$ and type II error $\beta_{\phi}=1-\mathbb{E}_{P_{1}}[\phi]$. 
A randomized mechanism $\mathcal{M}$ is of $\mu$-GDP if 
$$T(\mathcal{M}(D_{1}),\mathcal{M}(D_{2}))\geq T(\mathcal{N}(0,1),\mathcal{N}(\mu,1))$$ 
for any neighboring datasets $D_{1}, D_{2}$ differing by one record,
\end{definition}

The Gaussian mechanism can be applied to achieve $\mu$-GDP guarantees when releasing a statistic or a query result from a dataset.
\begin{definition}[Gaussian Mechanism of $\mu$-GDP]
\citep{dong2019gaussian}
For a given query $f$ and a dataset $D$, a Gaussian mechanism $\mathcal{M}(D)$ satisfying $\mu$-GDP is defined as 
$$f^*_{\mathcal{M}}(D)\sim\mathcal{N}(f(D),\Delta_{f}^{2}/\mu^{2}),$$
where $\Delta_{f} = \max\limits_{D_1, D_2} \| f(D_1)-f(D_2) \|_1$ for any pair of neighboring datasets $(D_{1}, D_{2})$.
\end{definition}
\citet{dong2019gaussian} also presents a noisy stochastic gradient descent (SGD) algorithm as a composition of  repeated  Gaussian mechanisms application with sub-sampling in
If we denote the number of iterations of the SGD algorithm by $T$, the sub-sampling rate by $r$, and the scale of noise added to the gradient at each iteration by $\sigma$, then the total privacy loss under GDP over $T$ iterations is $\mu=r\sqrt{T(e^{1/\sigma^{2}}-1)}$.

There is  a duality between $(\epsilon,\delta)$-DP and $\mu$-GDP~\citep{dong2019gaussian}, which helps to translate the privacy loss between the two DP frameworks. If a mechanism is of $\mu$-GDP, then it also satisfies $(\epsilon,\delta(\epsilon))$-DP, where
$\delta(\epsilon)=\Phi(-\frac{\epsilon}{\mu}+\frac{\mu}{2})-e^{\epsilon}\Phi(-\frac{\epsilon}{\mu}-\frac{\mu}{2}))$, where $\Phi$ is the CDF of the standard normal distribution.

\section{Private-preserving Density Estimation and Synthetic Data Generation through DP-NF}\label{sec:DP-NF-DE}

In this section, we present differentially private density estimation via NF, which can be used to synthesize privacy-preserving surrogate datasets before their public release.
Users can then conduct statistical analysis or train machine learning algorithms on the synthetic data in the same way as if they had the original data.
In what follows, we first present the EHR data that motivates the work, then provide an algorithm to incorporate DP in NF for density estimation, apply the algorithm to generate private EHR data, and finally examine the utility of the surrogate data via downstream learning tasks including classification, regression, and VI.

\subsection{Description of the EHR Dataset}\label{sec:EHR}

\begin{figure}[!htb]
\includegraphics[width=\linewidth]{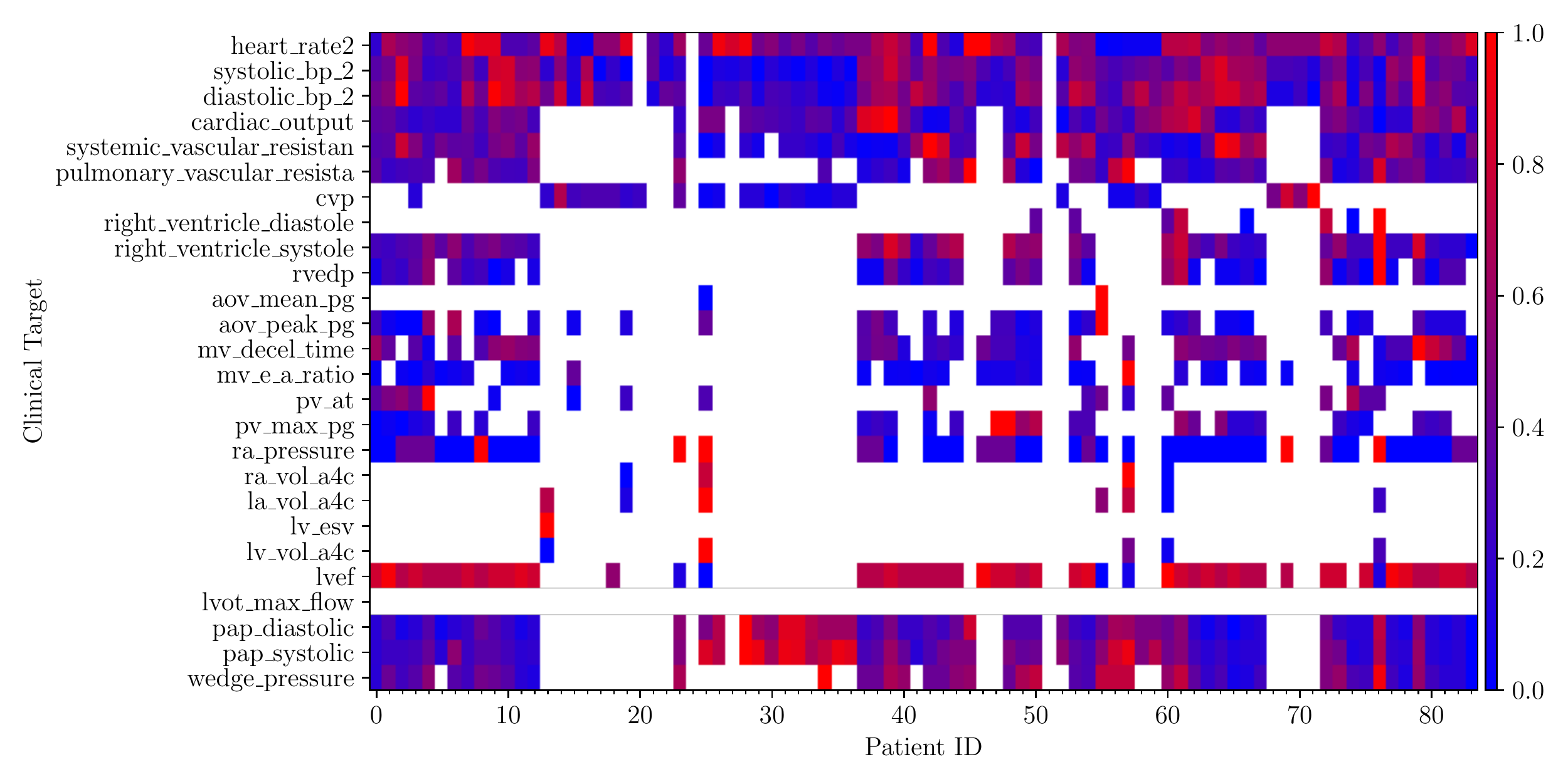}\vspace{-12pt}
\caption{Missing data pattern in EHR dataset. Each row in the above heat map is normalized to a $[0,1]$ range to highlight the relative magnitude of the clinical target across patients.}\vspace{-12pt}
\label{fig:data}
\end{figure}

The EHR dataset contains anonymized clinical measurements from 82 adult patients. The data were collected in the context of a research project funded by Google through its ATAP initiative, focusing on modeling noninvasive measurements of cardiovascular dynamics. Table~\ref{tab:PatientData} lists the 26 clinical attributes present in the dataset with the summary statistics. Figure~\ref{fig:data} shows the missing data pattern. Each row of the heat map was normalized to a zero to one range, to highlight the relative magnitude of each clinical target. Each patient has zero to nineteen clinical attributes. While there is not a single patient that has all 26 measurements, at least three clinical attributes (heart rate, diastolic blood pressure, and systolic blood pressure) are available for all but one patient. 
\begin{table}[!htb]
\begin{center}\vspace{-3pt}
\caption{Summary Statistics on the EHR dataset containing 26 clinical measurements.}\label{tab:PatientData}\vspace{-3pt}
\resizebox{\textwidth}{!}{
\begin{tabular}{l l p{7cm} p{2cm} p{1cm} p{1cm} p{1cm} p{1cm}}
\toprule
{\bf } & {\bf Attribute} & {\bf Description} & {\bf Unit} & {\bf SD$^*$} & {
\bf n$^\#$} & {\bf Min} & {\bf Max}\\
\midrule
1 & heart\_rate2 & Heart rate & bpm & 3.0 & 81 & 69.0 & 105.0\\
2 & systolic\_bp\_2 & Systolic blood pressure & mmHg & 1.5 & 81 & 83.0 & 180.0\\
3 & diastolic\_bp\_2 & Diastolic blood pressure & mmHg & 1.5 & 81 & 49.0 & 102.0\\
4 & cardiac\_output & Cardiac output & L/min & 0.2 & 65 & 3.3 & 10.9\\
5 & systemic\_vascular\_resistan & Systemic vascular resistance & dynes$\cdot$s$\cdot$cm$^{-5}$ & 50.0 & 64 & 668.6 & 2046.8\\
6 & pulmonary\_vascular\_resista & Pulmonary vascular resistance & dynes$\cdot$s$\cdot$cm$^{-5}$ & 5.0 & 50 & 5.0 & 245.8\\
7 & cvp & Central venous pressure & mmHg & 0.5 & 30 & 2.0 & 28.0\\
8 & right\_ventricle\_diastole & Right ventricle diastolic pressure & mmHg & 1.0 & 11 & 2.0 & 10.0\\
9 & right\_ventricle\_systole & Right ventricle systolic pressure & mmHg & 1.0 & 46 & 12.0 & 45.0\\
10 & rvedp & Right ventricle EDP & mmHg & 1.0 & 46 & 1.0 & 20.0\\
11 & aov\_mean\_pg & Average PG across aortic valve & mmHg & 0.5 & 2 & 5.0 & 10.0\\
12 & aov\_peak\_pg & Peak PG across aortic valve & mmHg & 0.5 & 37 & 3.0 & 18.0\\
13 & mv\_decel\_time & Mitral valve deceleration time & ms & 6.0 & 41 & 88.0 & 289.0\\
14 & mv\_e\_a\_ratio & Mitral valve E/A ratio & - & 0.2 & 39 & 0.7 & 20.9\\
15 & pv\_at & Pulmonary valve acceleration time & ms & 6.0 & 18 & 53.0 & 190.0\\
16 & pv\_max\_pg & Peak PG across pulmonary valve & mmHg & 0.5 & 31 & 1.0 & 18.0\\
17 & ra\_pressure & Mean right atrial pressure & mmHg & 0.5 & 50 & 3.0 & 15.0\\
18 & ra\_vol\_a4c & Right atrial volume & mL & 3.0 & 4 & 57.0 & 173.0\\
19 & la\_vol\_a4c & Left atrial volume & mL & 3.0 & 7 & 69.0 & 179.0\\
20 & lv\_esv & Left ventricular end-systolic volume  & mL & 10.0 & 1 & 32.0 & 32.0\\
21 & lv\_vol\_a4c & Left ventricular volume & mL & 20.0 & 5 & 59.0 & 642.0\\
22 & lvef & Left ventricular ejection fraction & - & 2.0 & 53 & 12.5 & 75.0\\
23 & lvot\_max\_flow & Peak flow velocity across LVOT & cm/s & - & 0 & - & -\\
24 & pap\_diastolic & Diastolic PAP & mmHg & 1.0 & 65 & 5.0 & 36.0\\
25 & pap\_systolic & Systolic PAP & mmHg & 1.0 & 65 & 10.0 & 60.0\\
26 & wedge\_pressure & Pulmonary wedge pressure & mmHg & 1.0 & 50 & 2.0 & 26.0\\
\hline
\end{tabular}}
\resizebox{\textwidth}{!}{
\begin{tabular}{l}
$^*$standard deviation; determined based on a preliminary literature review~\citep{yared2011pulmonary,gordon1983reproducibility,maceira2006normalized}.\\
$^\#$Number of patients with observed measurements on the respective attribute.\\
\bottomrule
\end{tabular}}\vspace{-18pt}
\end{center}
\end{table}
 
This dataset focuses on cases of group II pulmonary hypertension (i.e., hypertensive/not hypertensive), where a reversible increase of PVR follows an increase in left ventricular filling pressures\footnote{We do not consider pulmonary arterial hypertension (group I pulmonary hypertension) nor make any claim of differentiating group I and II pulmonary hypertension. Other criteria have been proposed to classify hypertension due to left ventricular diastolic dysfunction~\citep[see, e.g.,][]{simonneau2013updated}.}.
A often used binary classification  in practice is  a criterion based on two clinical measurements, the diastolic pulmonary artery pressure ($p_{p,d}$) and the systolic artery pressure ($p_{p,s}$), i.e.
\begin{equation}\label{eqn:hyper}
    {\rm hypertensive}=
    \begin{cases}
        1, & \text{if}\quad p_{p,m} = \frac{2}{3}\cdot p_{p,d}+\frac{1}{3}\cdot p_{p,s}>20\text{ mmHg}\\
        0, & \text{if}\quad p_{p,m} = \frac{2}{3}\cdot p_{p,d}+\frac{1}{3}\cdot p_{p,s}\leq20\text{ mmHg.}
    \end{cases}
\end{equation}

Medical information, including EHR data, is highly sensitive and may reveal patients' medical conditions and history, among others, if not properly protected for privacy. 
Sharing the data is not an option due to privacy concerns. Generating and releasing synthetic data is proposed as an alternative and is actively explored by researchers and in practice~\citep{shared, rankin2020reliability, dash2019privacy,yale2020generation,benaim2020analyzing}.

\subsection{Privacy-preserving Synthetic Data Generation through DP-NF}\label{sec:private_surrogate}

Figure~\ref{fig:data} shows that the EHR dataset contains a considerable portion of missing values. Before synthetic data generation, we first imputed the missing values using the \texttt{MICE} package (v3.13.0) in R.
The \texttt{MICE} package implements multivariate imputation by chained equations~\citep{van2011mice}, which draws, in an alternated fashion (as in a Gibbs' sampler), imputed values for the missing attribute $X_{\bigcdot,j}$ from the conditional distribution $p(X_{\bigcdot,j}|\bs\theta_j,\bm{X}_{\bigcdot,-j})$ and parameters $\bs\theta_j$ from its posterior distribution $p(\bs\theta_j|\bm{X})$ for $j=1,\ldots,p$. $\bm{X}_{\bigcdot,-j}$ contains all variables in the data except for $X_{\bigcdot,j}$.

In the EHR data, seven attributes (cvp, aov\_mean\_pg, ra\_vol\_a4c, la\_vol\_a4c, lv\_esc, lv\_vol\_a4c, and lvot\_max\_flow)  have large fractions of missing values that range from 63\%  to 100\%. 
These large fractions of missing data turned out to be problematic for MICE. For that reason, these seven attributes were not considered in the imputation and any subsequent analysis.
We obtained 5 sets of imputed data and then split each of the 5 imputed datasets into a training set and a testing set with 64 and 20 patients, respectively.
 
We used DP-NF for density estimation in each of the 5 imputed datasets using Algorithm~\ref{alg:DPNF_density}. 
After computing the privacy-preserving density $p^{*}(\bm{x})$ (we use $^*$ to represent private or sanitized quantities), we can generate synthetic data $\bm{x}^{*}$ by first sampling from the base distribution $\bm{z}_{0}\sim q_0$ and then use the learned NF map $F_{\lambda^{*}}(\bm{z}_{0})$ to generate realizations from $p^{*}$. 
Though the algorithm is presented using the SGD framework, it can easily be extended for other gradient-based optimization such as RMSprop~\citep{tieleman2012lecture} and Adam~\citep{kingma2014adam}, by replacing the update ($\lambda^{*(t+1)} = \lambda^{*(t)} - \eta\,g^{*(t)}$) with their respective parameter update paradigms.
\begin{algorithm}[H]
\caption{DP-NF for Density Estimation}\label{alg:DPNF_density}
\begin{algorithmic}
\REQUIRE{data set $\bm{x}=\{\bm{x}_{1},\ldots,\bm{x}_{n}\}$, initial parameter values $\lambda^{(0)}$ of NF $F_{\lambda}(\cdot)$ with base distribution $q_0$, learning rate $\eta$, DP noise scale $\sigma$, subsampling rate $r$, clipping constant on gradient $C$, number of iterations $T$.}
\ENSURE{privacy-preserving estimated density $p^*(\bm{x})$}
\FOR{$t=0,\cdots,T$}
\STATE{sub-sample $\bm x^{(t)}$ from $\bm{x}$ with rate $r$; let $b^{(t)}$ be the size of $\bm x^{(t)}$}
\FOR{$\bm x_i\in \bm  x^{(t)}$}
\STATE{let $\bm z_i\!=\!F_{\lambda}^{-1}(\bm x_i)$}
\STATE{obtain $l_i=-\log q_0(\bm z_i) - \log |\partial F_{\lambda} / \partial \bm z_i|$}
\STATE{compute  $\bm g_i^{(t)} = \nabla_{\lambda}l_i$}
\STATE{$\bm g_i^{(t)} \leftarrow \bm g_i^{(t)}/\max(1, \|\bm g_i^{(t)}\|_2/C)$}
\ENDFOR
\STATE{obtain sanitized gradient $\bm g^{*(t)} =\left(\sum_{i=1}^{b^{(t)}} \bm g_i^{(t)} + \mathcal{N}(\bm 0,\sigma^{2}C^2\bm I)\right)/b^{(t)}$}
\STATE{$\lambda^{*(t+1)} = \lambda^{*(t)} - \eta\,g^{*(t)}$}
\ENDFOR
\end{algorithmic}
\end{algorithm}

In the application of Algorithm \ref{alg:DPNF_density} to the EHR data, we set $r= 0.5, \eta= 2\times10^{-5}, T= 8000, C= 10$. We examined 4 privacy budgets $\mu$ in the setting of $\mu$-GDP, i.e., $\mu=(6.10, 3.92, 2.45, 1.49$)\footnote{We also attempted smaller $\mu$ (e.g. $\mu\le 1$), but the DP-NF algorithm did not converge due to a large amount of noise injected to the gradients.}, corresponding to $\sigma= (7.36, 11.44, 18.28, 29.93)$ and $\epsilon = (32, 16, 8,4)$ if $\delta=0.01$ in the $(\epsilon,\delta)$-DP setting. 
The NF architecture we used has 15 MADE layers, each consisting of a fully connected neural network having 1 hidden layer with 200 neurons, and one batch normalization layer following each MADE. 
For each of the 5 imputed training sets, we run DP-NF 10 times to obtain 10 privacy-preserving densities; from each estimated density, 10 synthetic datasets were generated,  leading to a total of 500 synthetic data sets\footnote{We generated multiple data sets to examine the stability of the results in the downstream learning tasks in Sections~\ref{sec:prediction} and \ref{sec:VI}. In practice, one may release multiple synthetic data to take into account the uncertainty around sanitization and synthesis, and then combine the inferences from multiple sets using the rule in~\citet{liu2016model}. 
Correspondingly, the overall privacy budget needs to split across multiple sets when generating privacy-preserving synthetic data. 
For machine learning tasks based on synthetic data, releasing a single synthetic dataset may be sufficient.}.
Each synthetic data contains the same number of subjects (82) as in the original dataset.

We use the generated synthetic data for three subsequent learning and inferential tasks. 
The first two focus on detecting pulmonary hypertension and predicting the mean pulmonary arterial pressure from the testing data (Section~\ref{sec:prediction}), while the last conducts VI for the parameters of a physics-based model (Section~\ref{sec:VI}).

\subsection{Classification and Regression based on Privacy-preserving Synthetic EHR Data}\label{sec:prediction}

The goal of this learning task is to detect pulmonary hypertension, defined as a mean pulmonary artery pressure $p_{p,m}$ greater than 20 mmHg in Eqn.~\eqref{eqn:hyper}.
Toward that end, we train a binary classifier (hypertensive/not hypertensive) using support vector machine~\citep[SVM,][]{cortes1995support} and a numerical estimate of the mean pulmonary artery pressure using a random forest~\citep[see, e.g.,][]{breiman2001random}.
We evaluated the classification and regression testing accuracy of these two procedures trained on the privacy-preserving synthetic data.
SVM outputs a probability of having pulmonary hypertension for each case. Instead of using a cutoff of 0.5, we adopted the Fowlkes–Mallows (FM) index~\citep{fowlkes1983method} to find an optimal cutoff. 
The FM index is defined as $\sqrt{\frac{\text{TP}}{\text{TP+FP}}\cdot\frac{\text{TP}}{\text{TP+FN}}}$, where TP is the number of true positives, FP is the number of false positives, and FN is the number of false negatives.
The cutoff that produces the maximum FM index was used.

Tables~\ref{tab: SVM1} and~\ref{tab: RF1} summarize the prediction results over the 500 synthetic datasets. Both tables present two sets of results, one based on SVM or random forest trained on the raw generated synthetic data, the other based on SVM or random forest trained on post-processed synthetic data by projecting an out-of-bound synthetic data point back to either its lower or upper bound.
The reason for the post-processing is that the attributes in the EHR data are clinical attributes and they are naturally bounded. 
For example, the attribute ``peak pressure gradient across pulmonary valve'' relates to pulmonary valve stenosis; values outside $[0,50]$ would be regarded as physiologically impossible for our EHR dataset which does not contain patients with pulmonary valve stenosis. 
Therefore, if a synthetic value on ``peak PG across pulmonary valve'' is $<0$ mmHg, it is set at 0 and, is set at 50 mmHg if found $>50$ mmHg.

\begin{table}[!htb]
\centering
\caption{Privacy-preserving mean classification accuracy rate (SD) via SVM based on synthetic data generated via DP-NF}
\label{tab: SVM1}
\resizebox{0.95\textwidth}{!}{
\begin{tabular}{cccccccc}
  \toprule
Truncation$^*$ &  Imputation & Original & $\mu=\infty$ & $\mu=6.10$ & $\mu=3.92$ & $\mu=2.45$ & $\mu=1.49$ \\ 
\midrule
& 1 & 1.00 & 0.95 (0.07) & 0.91 (0.08) & 0.82 (0.24) & 0.81 (0.15) & 0.87 (0.09) \\ 
& 2 & 1.00 & 0.88 (0.24) & 0.93 (0.06) & 0.86 (0.10) & 0.88 (0.12) & 0.83 (0.11) \\ 
No & 3 & 1.00 & 0.92 (0.09) & 0.91 (0.06) & 0.90 (0.08) & 0.88 (0.10) & 0.72 (0.38) \\
& 4 & 0.95 & 0.89 (0.10) & 0.88 (0.05) & 0.80 (0.09) & 0.76 (0.22) & 0.86 (0.12) \\ 
& 5 & 0.95 & 0.85 (0.20) & 0.90 (0.07) & 0.88 (0.17) & 0.92 (0.17) & 0.80 (0.10) \\ 
\cline{2-8}
& mean & 0.98 & 0.90 (0.16) & 0.91 (0.07) & 0.85 (0.15) & 0.85 (0.20) & 0.81 (0.20)\\ 
\hline
& 1 & 1.00 & 0.92 (0.08) & 0.87 (0.09) & 0.79 (0.24) & 0.80 (0.15) & 0.85 (0.09) \\
& 2 & 1.00 & 0.83 (0.26) & 0.89 (0.07) & 0.86 (0.09) & 0.86 (0.11) & 0.83 (0.10) \\
Yes & 3 & 1.00 & 0.90 (0.11) & 0.90 (0.07) & 0.90 (0.08) & 0.88 (0.10) & 0.70 (0.37) \\
& 4 & 0.95 & 0.85 (0.16) & 0.87 (0.06) & 0.80 (0.08) & 0.74 (0.33) & 0.82 (0.12) \\
& 5 & 0.95 & 0.77 (0.28) & 0.85 (0.08) & 0.85 (0.17) & 0.88 (0.17) & 0.80 (0.10) \\
\cline{2-8}
& mean & 0.98 & 0.85 (0.20) & 0.88 (0.08) & 0.84 (0.15) & 0.83 (0.20) & 0.80 (0.20)\\ 
\hline
\end{tabular}}
\resizebox{0.95\textwidth}{!}{
\begin{tabular}{l}
$^*$ Imputed and synthetic data on an attribute may exceed the physiologically feasible bounds. The results in  \\``Truncation = Yes'' are obtained based on post-processed imputed and synthetic data by setting the out-of- \\ 
bound values at the closest bound. \\
\bottomrule
\end{tabular}}
\end{table}

\begin{table}[!htb]
\centering
\caption{Privacy-preserving prediction mean squared error  (SD) via random forest based on synthetic data generated via DP-NF}\label{tab: RF1}
\resizebox{1\textwidth}{!}{
\begin{tabular}{cccccccc}
\toprule
Truncation$^*$ &  Imputation & Original & $\mu=\infty$ & $\mu=6.10$ & $\mu=3.92$ & $\mu=2.45$ & $\mu=1.49$ \\ 
\midrule
 & 1 & 20.63 & 40.30 (23.37) & 54.71 (20.20) & 62.45 (20.55) & 83.44 (42.54) & 70.39 (30.52)\\ 
 & 2 & 18.02 & 35.70 (19.87) & 48.08 (18.10) & 60.06 (22.89) & 62.44 (33.44) & 80.30 (44.03)\\ 
No & 3 & 11.50 & 71.76 (62.20) & 36.53 (7.39) & 47.02 (23.01) & 72.22 (28.51) & 71.36 (30.02)\\ 
 & 4 & 18.87 & 32.92 (27.36) & 44.15 (12.19) & 63.09 (19.02) & 46.81 (26.87) & 60.52 (17.59)\\ 
 & 5 & 20.66 & 81.45 (59.57) & 53.98 (17.75) & 59.53 (29.41) & 41.09 (18.28) & 122.91 (41.70)\\ 
\cline{2-8}
& mean & 17.94 & 52.52 (47.18) & 47.40 (17.17) & 58.41 (23.86) & 61.39 (34.49) & 80.10 (39.92)\\
\hline
 & 1 & 21.49 & 47.50 (27.68) & 72.30 (23.98) & 79.00 (26.64) & 95.76 (43.68)& 75.85 (28.33)\\ 
 & 2 & 18.54 & 47.08 (24.01) & 56.77 (15.78) & 68.91 (22.45) & 73.43 (36.21) & 88.21 (42.44)\\ 
Yes & 3 & 12.12 & 68.25 (55.06) & 47.09 (12.74) & 55.26 (22.01) & 80.08 (29.44) & 85.45 (29.94)\\ 
 & 4 & 18.06 & 35.55 (30.28) & 55.24 (13.12) & 71.30 (24.08) & 51.26 (26.73) & 66.17 (22.38)\\ 
 & 5 & 19.79 & 78.79 (53.02) & 63.31 (18.31) & 68.13 (31.79) & 53.75 (21.02) & 112.54 (40.61)\\
\cline{2-8}
& mean & 18.00 & 55.91 (42.61) & 58.94 (19.23) & 68.52 (26.77) & 71.91 (35.70) & 84.91 (36.52)\\ 
\bottomrule
\end{tabular}}
\resizebox{1\textwidth}{!}{
\begin{tabular}{l}
$^*$ Imputed and synthetic data on an attribute may exceed the physiologically feasible bounds. The results in  \\``Truncation = Yes'' are obtained based on post-processed imputed and synthetic data by setting the out-of- \\ 
bound values at the closest bound. \\
\bottomrule
\end{tabular}}\vspace{-9pt}
\end{table}
Overall, as expected, the higher the privacy loss (larger $\mu$), the higher the accuracy rates (Table~\ref{tab: SVM1}) and the smaller the prediction MSE (Table~\ref{tab: RF1}).  
The post-processing truncation needed to enforce bounds for each clinical measurement satisfies the practical data constraints but negatively affects the prediction results; the accuracy rates decrease in Table~\ref{tab: SVM1} and the prediction MSE increases in Table~\ref{tab: RF1} with truncation vs. without truncation, regardless of the preferred privacy budget; but the trend over $\mu$ is similar to that before the truncation.

\subsection{VI of Parameters in Physics-based model using Privacy-preserving Synthetic  EHR data} \label{sec:VI}

\subsubsection{the CVSim-6 physiologic model}

\begin{figure}[!htb]
\centering
\begin{subfigure}{0.32\textwidth}
\centering
\includegraphics[width=\textwidth]{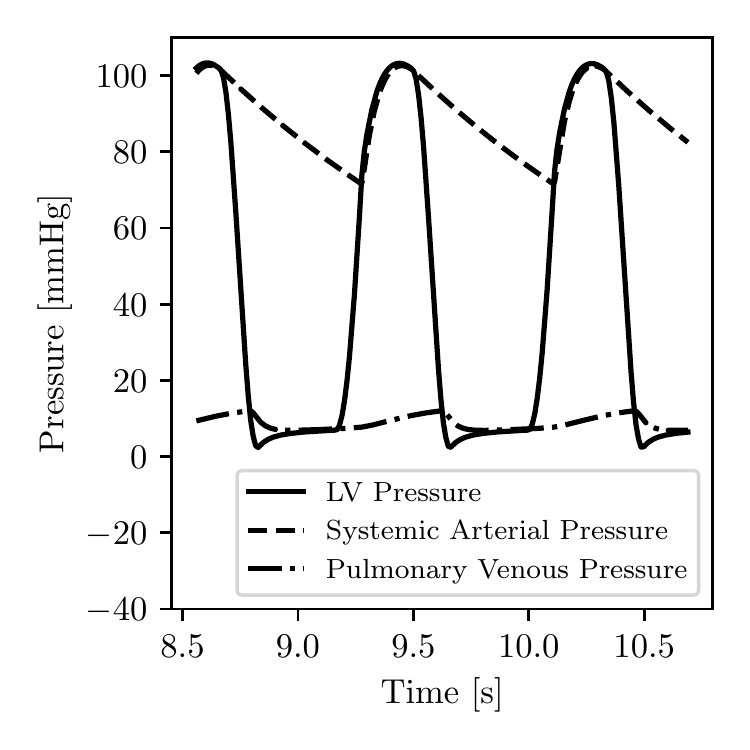}\vspace{-12pt}
\caption{LV pressures.}
\end{subfigure}
\begin{subfigure}{0.32\textwidth}
\centering
\includegraphics[width=\textwidth]{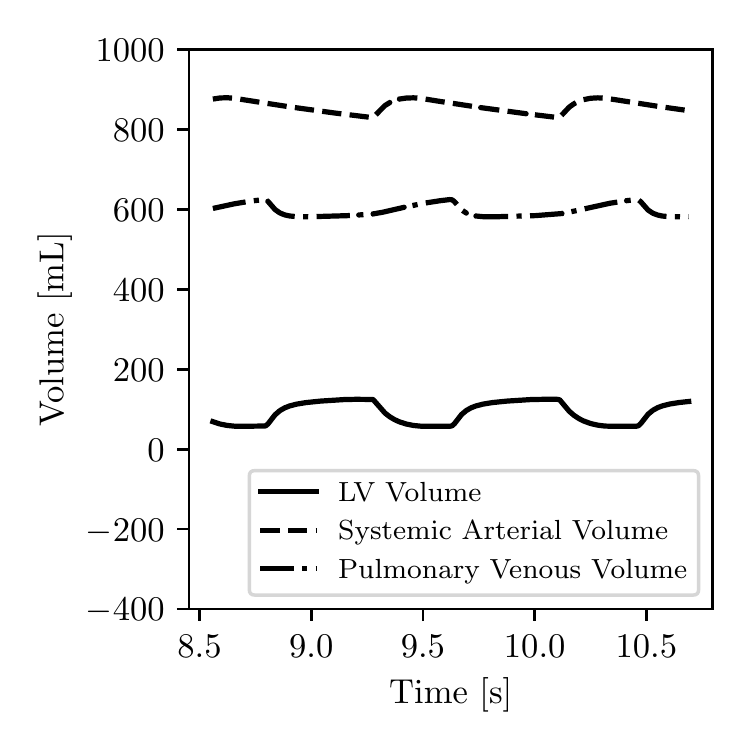}\vspace{-12pt}
\caption{LV volumes.}
\end{subfigure}
\begin{subfigure}{0.32\textwidth}
\centering
\includegraphics[width=\textwidth]{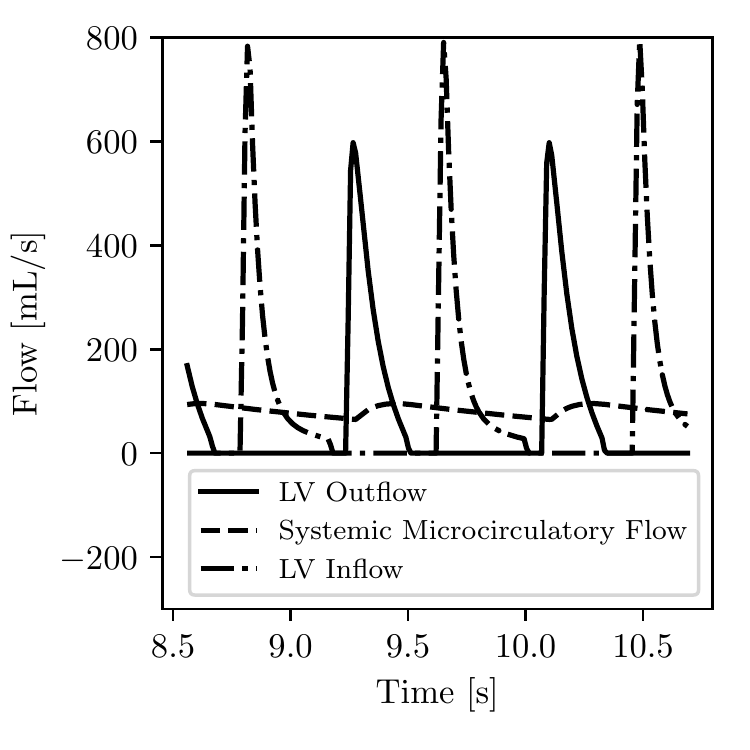}\vspace{-12pt}
\caption{LV flows.}
\end{subfigure}

\begin{subfigure}{0.32\textwidth}
\centering
\includegraphics[width=\textwidth]{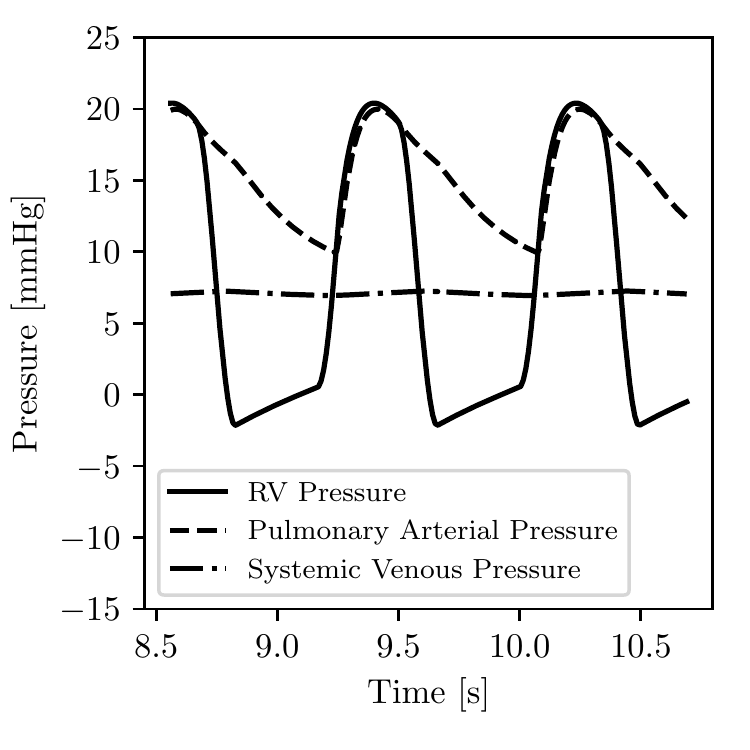}\vspace{-12pt}
\caption{RV pressures.}
\end{subfigure}
\begin{subfigure}{0.32\textwidth}
\centering
\includegraphics[width=\textwidth]{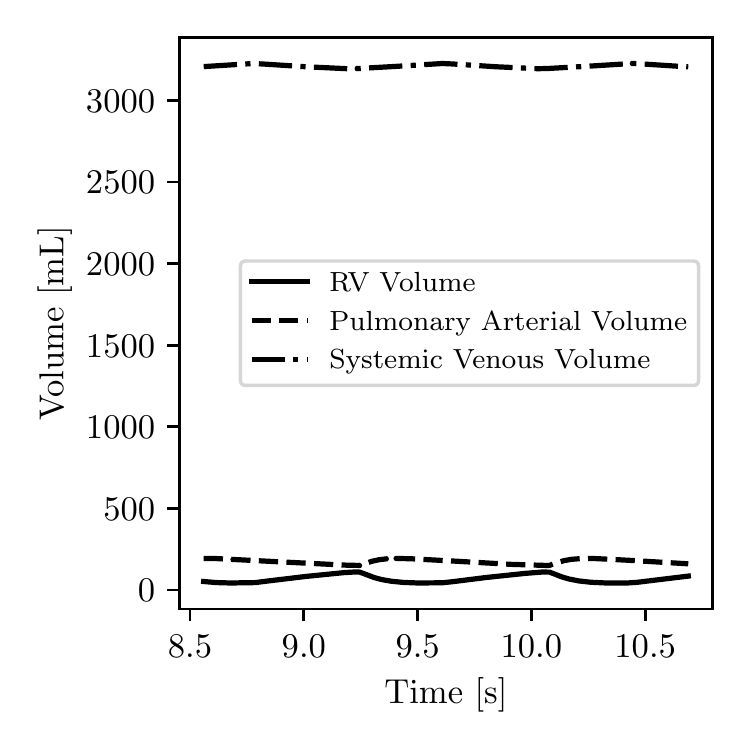}\vspace{-12pt}
\caption{RV volumes.}
\end{subfigure}
\begin{subfigure}{0.32\textwidth}
\centering
\includegraphics[width=\textwidth]{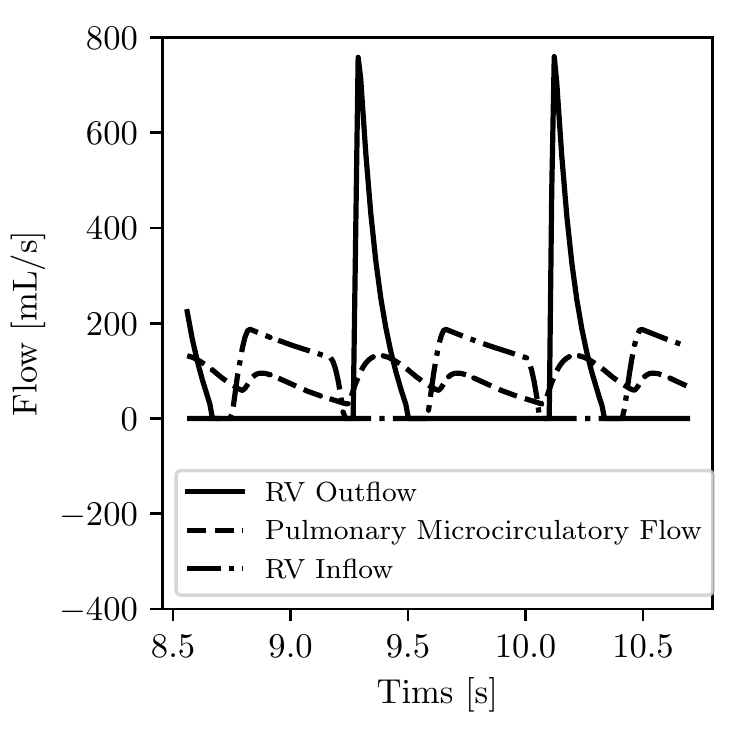}\vspace{-12pt}
\caption{RV flows.}
\end{subfigure}
\caption{Time histories of state and auxiliary variables from CVSim-6 model, when using the default input parameters listed in Table~\ref{tab:cvsim_def}~\citep[see Figure 4-4 in][]{davis1991teaching}.}
\label{fig:cvsim}
\end{figure}

The evolution of blood pressure, flow, and volume in the circulatory system of an adult subject is simulated, in this study, using CVSim-6~\citep{davis1991teaching,heldt2010cvsim}, a lumped parameter hemodynamic model with 6-compartments available through the PhysioNet repository~\citep{goldberger2000physiobank}. A custom Python/Cython implementation can be found at \url{https://github.com/desResLab/supplMatHarrod20}. 
The model includes compartments for the left heart, right heart, systemic arteries, systemic veins, pulmonary arteries and pulmonary veins.
Arterial and venous compartments consist of RC circuits, while heart chambers are simulated using a pressure generator, two unidirectional diodes and a resistor to account for the pressure loss produced at the valves. 
All elements are considered linear, disregarding collapsibility in veins associated with negative pressures. 
Inertial effects are also assumed negligible (the model does not contain inductors) and a two-chamber heart is considered, disregarding any contribution from the atria.
In addition, interaction is only considered between adjacent compartments. Thus, unlike other model formulations in the literature~\citep[see, e.g., CircAdapt, see][]{arts2005adaptation} no mechanical interaction is considered between the two ventricular chambers through the septal wall.
CVSim-6 consists of a system of six differential equations (one per compartment). The flows between the compartments, under the assumption of nonlinear unidirectional valves (without regurgitation), are expressed as~\cite[see][Sections 4.2.1 and 4.2.2]{davis1991teaching}
%
\begin{equation}
\begin{tabular}{cc}
$
q_{li} = 
\begin{cases}
(P_{pv} - P_{l})/R_{li} & \text{if}\,\,P_{pv} > P_{l}\\
0 & \text{otherwise}
\end{cases}$ & 
$q_{lo} = 
\begin{cases}
(P_{l} - P_{a})/R_{lo} & \text{if}\,\,P_{l} > P_{a}\\
0 & \text{otherwise}
\end{cases}$\\
\\
$q_{ri} = 
\begin{cases}
(P_{v} - P_{r})/R_{ri} & \text{if}\,\,P_{v} > P_{r}\\
0 & \text{otherwise}
\end{cases}$
&
$q_{ro} = 
\begin{cases}
(P_{r} - P_{pa})/R_{ro} & \text{if}\,\,P_{r} > P_{pa}\\
0 & \text{otherwise}
\end{cases}$\\
\\
$q_{a} = (P_{a} - P_{v})/R_{a}$ & 
$q_{pv} = (P_{pa} - P_{pv})/R_{pv}$\\
\end{tabular},
\end{equation}
leading to the following differential equations for RC elements with fixed and time-varying capacitance 
\begin{equation}
\begin{split}
\dfrac{dP_{l}}{dt} &= \dfrac{q_{li}-q_{lo}- (P_{l}-P_{th})\,dC_{l}(t)/dt}{C_{l}(t)},\quad \dfrac{dP_{a}}{dt} = \dfrac{q_{lo}-q_{a}}{C_{a}},\\
\dfrac{dP_{v}}{dt} &= \dfrac{q_{a}-q_{ri}}{C_{v}},\quad \dfrac{dP_{r}}{dt} = \dfrac{q_{ri}-q_{ro}- (P_{r}-P_{th})\,dC_{r}(t)/dt}{C_{r}(t)},\\
\dfrac{dP_{pa}}{dt} &= \dfrac{q_{ro}-q_{pv}}{C_{pa}},\quad\,\,\dfrac{dP_{pv}}{dt} = \dfrac{q_{pv}-q_{li}}{C_{pv}}.\\
\end{split}
\end{equation}
Initial conditions are determined from a system of linear equilibrium equations on the stressed volume,
\begin{align}
C_{l,dias}\,(P_{l,dias}-P_{th}) - C_{l,sys}\,(P_{l,sys}-P_{th}) = &\,C_{r,dias}\,(P_{r,dias}-P_{th}) - C_{r,sys}\,(P_{r,sys}-P_{th})\notag\\
= &\,T_{sys}\,\dfrac{P_{l,sys}-P_{a}}{R_{lo}} = T_{tot}\,\dfrac{P_{a}-P_{v}}{R_{a}}\\
= &\,T_{dias}\,\dfrac{P_{v}-P_{r,dias}}{R_{v}} = T_{sys}\,\dfrac{P_{r,sys}-P_{pa}}{R_{ro}}\notag\\
= &\,T_{tot}\,\dfrac{P_{pa}-P_{pv}}{R_{pv}} = T_{dias}\,\dfrac{P_{pv}-P_{l,dias}}{R_{li}}\notag\\
V_{tot} - V_{0,tot} = &\,C_{l,dias}(P_{l,dias}-P_{th}) + C_{a}
(P_{a}-\frac{1}{3}P_{th}) + \notag\\
&\,C_{v}\,P_{v} + C_{r}(P_{r} - P_{th}) + C_{pa}(P_{pa} - P_{th})+\notag\\ 
&\,C_{pv}(P_{pv}-P_{th}).\notag
\end{align}
The acronyms used in the equations above are defined in Table~\ref{tab:acro} in the Appendix.
Default model parameters are reported in Table~\ref{tab:cvsim_def} and pressure, volume, and flow rate time histories produced by CVSim-6, 
when such parameters are used as inputs, are shown in Figure~\ref{fig:cvsim}. 

\begin{table}[!htb]
\caption{Default parameters for CVSim-6 model (see Table 4.1 in ~\cite{davis1991teaching})}\label{tab:cvsim_def}
\resizebox{\textwidth}{!}{  
\begin{tabular}{lcccc}
\toprule
{\bf Compartment} & {$R_{i}$ (mmHg s/mL)} & {$R_{o}$ (mmHg s/mL)} & { C (mL/mmHg)} & {$V_{0}$ (mL)}\\
\midrule
{\bf Left Heart} & 0.01 & 0.006 & 0.4-10 & 15\\
{\bf Systemic Arteries} & 0.006 & 1.00 & 1.6 & 715\\
{\bf Systemic Veins} & 1.00 & 0.05 & 100.0 & 2500.0\\
{\bf Right Heart} & 0.05 & 0.003 & 1.2-20 & 15\\
{\bf Pulmonary Arteries} & 0.003 & 0.08 & 4.3 & 90\\
{\bf Pulmonary Veins} & 0.08 & 0.01 & 8.4 & 490\\
\hline
\end{tabular}}
\resizebox{\textwidth}{!}{  
\begin{tabular}{lcccc}
{\scriptsize Systemic parameters: total blood volume = 5000 mL, heart rate = 72 bpm; thransthoracic pressure = -4.0 mmHg.}\\
\bottomrule
\end{tabular}}
\end{table}

The CVSim-6 model is compactly represented as $\bm{x} = \bm{f}(\bm{z})$, where $\bm{z}$ and $\bm{x}$ represent the model inputs (parameters, either unknown or fixed at constants) and outputs (observed), respectively. 
A subset consisting of 8 outputs was in the construction of  the likelihood function given the EHR data, including heart rate, pulmonary vascular resistance, central venous pressure, right ventricular diastolic pressure, right ventricular systolic pressure, right ventricle end-diastolic pressure, average pressure gradient across the aortic valve, and peak pressure gradient across the aortic valve. Such quantities are obtained by post-processing the time histories for the state and auxiliary variables shown in Figure~\ref{fig:cvsim}. 
In this study, we regard right ventricular resistance $R_{ro}$ and arterial capacitance $C_{a}$ as unknown parameters (inputs to the model), whereas the other parameters are fixed to their default values in Table~\ref{tab:cvsim_def}.

\subsubsection{VI based on privacy-preserving synthetic EHR data}\label{sec:DP-VI-syn-EHR}

To infer the CVSim-6 model parameters $R_{ro}$ and $C_{a}$ for  hypertensive patients, we construct the likelihood based on a subset of 43 synthetic hypertensive patients. We assume a uniform prior of the transformed parameters $(R', C')$ on $[-7, 7]^2$, 
\begin{equation}
(R_{ro}, C_{a}) = (\tanh(3/7 \cdot R')\cdot(H_{R} - L_{R})/2 + R_0, \tanh(3/7 \cdot C')\cdot(H_{C} - L_{C})/2 + C_0),
\end{equation}
where $R_{ro}\in[L_R, H_R]$, $C_{a}\in[L_C, H_C]$ are the original parameters with respective bounds, $R_0, C_0$ are default values, and $(R', C')$ are their transformed unbounded counterpart. 

We assume the  8-dimensional  output $\bm{x}_{i} = (x_{i,1},\ldots, x_{i,8})$, given the parameters $(R_{ro},C_{a})$, follows a multivariate Gaussian distribution. Assuming independence across the 8 outputs with known marginal variance for each output, the he log-likelihood of $(R_{ro},C_{a})$ is 
\begin{equation}
\ell(R_{ro},C_{a};\bm x) = -\frac{1}{2}\sum_{i=1}^n\,\left[\bm{x}_i - \bm{f}(R_{ro},C_{a})\right]^T\,\bm{\Sigma}^{-1}\,\left[\bm{x}_i - \bm{f}(R_{ro},C_{a})\right]^2 +\mbox{constant},
\end{equation}
where $\bm{f}(R_{ro},C_{a})$ represents the output of the CVSim-6 model with all but two input variables fixed at their default values and $\bm{\Sigma}$ is a diagonal matrix with marginal variances equal to (444.84, 145.60, 2.21, 45034.41, 475.59, 24.77, 84.06, 15.48) for the 8-dimensional output. 

The CVSim-6 model is computationally expensive. To reduce the computational cost of VI with input parameters $R_{ro}$ and $C_{a}$,  we trained offline a fully connected neural network surrogate.
The network contains 2 hidden layers with 64 and 32 neurons, respectively, and Tanh activations. We generated 100 realizations using Sobol' sampling in the ($R_{ro}$,$C_{a}$) plane and computed the corresponding outputs from the CVSim-6 model. 
We then trained the surrogate model based on 100 sets of inputs (2-dimensional) and outputs (8-dimensional) in 120,000 iterations with the RMSprop optimizer (learning rate 0.01, learning rate scheduler with exponential decay factor 0.9999) using a $\ell_2$ loss with $\ell_2$ regularization  (penalty 0.0001). 

We train the surrogate model and run NF to obtain posterior distribution on the transformed $(R', C')$ first, which are then transformed back to obtain the posterior distribution and VI of $R_{ro}$,$C_{a}$. 
VI was performed using a MAF architecture with 5 alternated MADE and batch normalization layers. Each MADE uses a fully connected neural network characterized by 1 hidden layer with 100 neurons. The activation functions within and between successive MADE layers are ReLU. 
The network is trained for 15,000 iterations, generating batches of 500 samples at each iteration. 
We use RMSProp as the optimizer with a learning rate of 0.01 and an exponential scheduler with a decay rate of 0.9999.  

VI results for $R_{ro}$ and $C_a$ based on the privacy-preserving synthetic data are presented in Table~\ref{tab:VI}, showing the posterior mean and posterior standard deviations of the two parameters, and the correlations between them across 5 different imputations.
In summary, the estimates based on privacy-preserving synthetic data are somewhat different from those based on the original data and there does not appear to be an obvious trend over $\mu$ in terms of how the former differs from the latter, except for the posterior correlation between $R_{ro}$ and $C_a$, which appears to get more negative as $\mu$, the privacy loss, decreases. 
\begin{table}[!htb]\vspace{-9pt}
\centering \caption{Privacy-preserving VI for CVSim-6 model parameters $R_{ro}$ and $C_a$  based on synthetic data generated via DP-NF}\label{tab:VI}
\resizebox{1\textwidth}{!}{
\begin{tabular}{ccccccc}
\toprule
parameter &  \makecell[c]{{posterior}\\{estimate}} &original & $\mu=6.10$ & $\mu=3.92$ & $\mu=2.45$ & $\mu=1.49$ \\ 
\midrule
$R_{ro}$	& mean	&\ 35.70 (\ 1.918)	&\ 43.83 (\ 2.798)	&\ 45.90 (\ 2.788)	&\ 47.52 (\ 4.400)	&\ 45.61 (\ 3.748)	\\
	& SD	&\ 1.452 (\ 0.038)	&\ 1.907 (\ 0.032)	&\ 2.168 (\ 0.341)	&\ 2.189 (\ 0.383)	&\ 2.144 (\ 0.165)	\\
\midrule
$C_a$	& mean ($\times10^{-4}$)	&\ 6.520 (\ 0.130)	&\ 8.569 (\ 1.169)	&\ 8.345 (\ 1.647)	&\ 7.972 (\ 0.967)	&\ 8.917 (\ 1.023)	\\
	& SD ($\times10^{-4}$)	&\ 0.347 (\ 0.013)	&\ 0.609 (\ 0.135)	&\ 0.666 (\ 0.362)	&\ 0.594 (\ 0.200)	&\ 0.724 (\ 0.192)	\\
\midrule
\multicolumn{2}{c}{$R_{ro},C_a$ correlation}	&-0.120 (\ 0.042)	&-0.123 (\ 0.192)	&-0.169 (\ 0.275)	&-0.213 (\ 0.244)	&-0.193 (\ 0.164)	\\
\bottomrule
\end{tabular}}\vspace{-12pt}
\end{table}

\begin{figure}[!htb]
\centering
\includegraphics[trim= 2.5cm 0cm 2.5cm 0.5cm,clip,width=\linewidth]{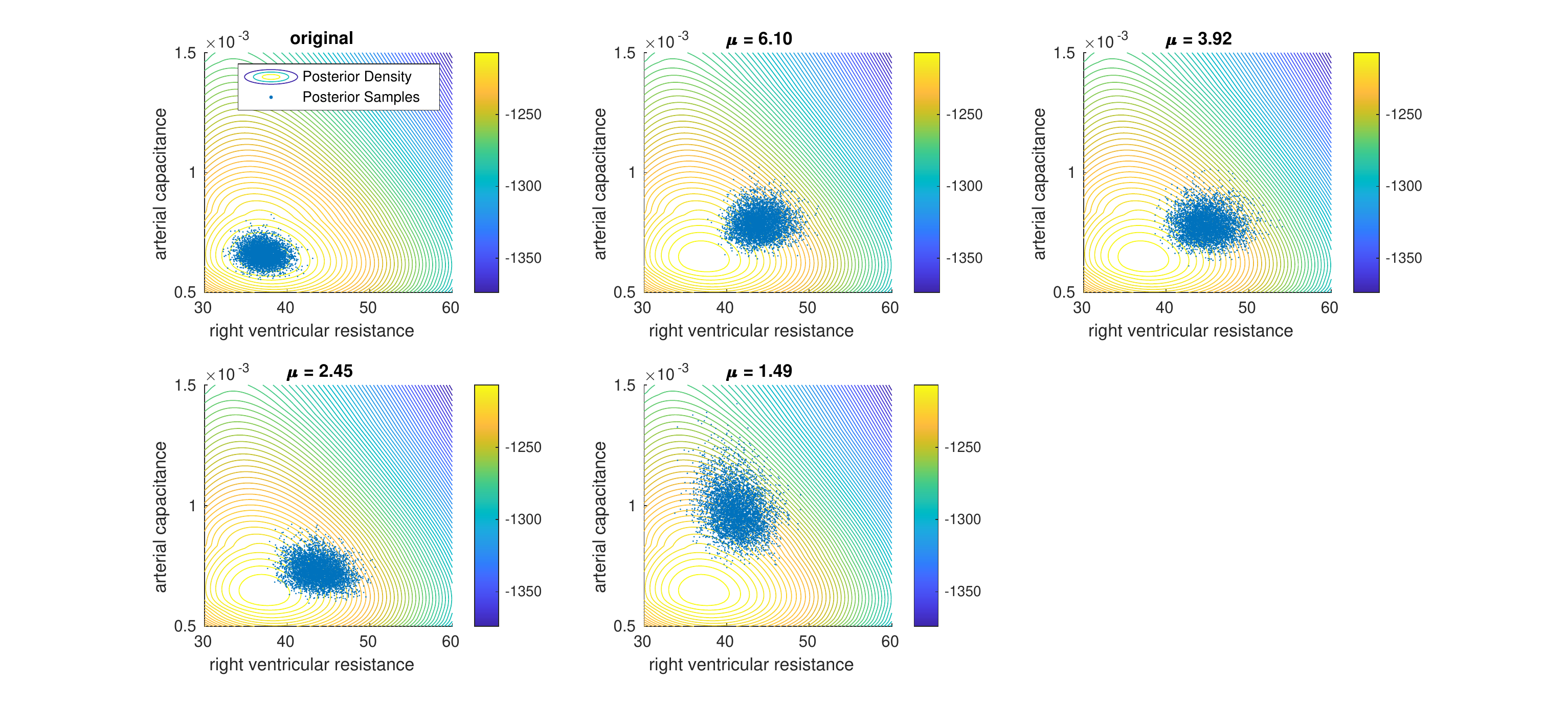}
\caption{Example of privacy-preserving posterior samples of $R_{ro}$ and $C_a$ through MAF based on one synthetic dataset. The contours are the densities of the true posterior distribution of $R_{ro}$ and $C_a$ given the true model and the original data.}\label{fig:VI}\vspace{-12pt}
\end{figure}

Figure~\ref{fig:VI} presents the scatter plots of the posterior samples on $R_{ro}$ and $C_a$ from the NF run on one private synthetic dataset. 
The observations are consistent with those in Table~\ref{tab:VI}. In general, the posterior samples generated from the NF constructed from the privacy-preserving synthetic data are located near the high-density regions of the posterior distribution based on the original data and true model, though not quite around the mode, and the mild negative correlation between the two parameters is also roughly retained for $\mu\ge 2.45$. 
There is more dispersion among the posterior samples at all $\mu$, especially at $\mu=1.49$, due to the noise injected to achieve DP guarantees, as expected.

\section{Privacy-Preserving VI via DP-NF}\label{sec:DP-NF-VI}
In this section, we examine the feasibility of privacy-preserving VI through sanitization of the optimization procedure employed by NF for VI, referred as the DP-NF-VI Procedure for short. Though NF can be used for both density estimation and VI and gradient-based optimization can be used to obtain the respective solutions in both cases, the loss functions in the two cases are different (likelihood in Eqn.~\eqref{eqn:MLE} vs. free energy bound in Eqn.~\eqref{eqn:KL}). 
In what follows, we first provide an algorithm for DP-NF-VI with DP guarantees for NF-VI and then examine the inferential utility of privacy-preserving VI via DP-NF in simulated data.

\subsection{The DP-NF-VI Procedure}
The steps of the DP-NF-VI Procedure is given in Algorithm~\ref{alg:DPNF_VI} 
Similar to Algorithm~\ref{alg:DPNF_density}, Algorithm~\ref{alg:DPNF_VI} is obtained from an SGD optimizer by clipping and perturbing the gradient with additive noise; and it can easily be extended to other gradient-based optimization procedures such as RMSprop or Adam  by replacing the parameter updating step ($\lambda^{*(t+1)} = \lambda^{*(t)} - \eta\,g^{*(t)}$) with their respective parameter update paradigms.
The differences between the two algorithms are in the loss function and the privacy-preserving output which are the free energy and the privacy-preserving variational density approximation $p^{*}(\bm z)$, respectively, in Algorithm~\ref{alg:DPNF_density}. 
We can draw samples from privacy-preserving samples of $\bm z$ from $p^{*}(\bm z)$ by first generating $\bm{z}_{0}\sim q_0$ and then applying the transformation $F_{\lambda^{*}}(\bm{z}_{0})$. 

\begin{algorithm}[H]
\caption{The DP-NF-VI Procedure}\label{alg:DPNF_VI}
\begin{algorithmic}
\REQUIRE{observed data $\bm{x}=\{\bm{x}_{1},\ldots,\bm{x}_{n}\}$,  target posterior distribution $p(\bm z | \bm x)$,  NF $F_{\lambda}(\bm z) = f_K \circ f_{K-1} \circ \cdots f_{1}(\bm z)$ with base distribution $q_0$, initial values $\lambda^{(0)}$, learning rate $\eta$, DP noise scale $\sigma$, sub-sampling rate $r$, gradient clipping constant $C$, number of iterations $T$, batch size $m$.}
\ENSURE{privacy-preserving variational distribution $p^{*}(\bm z)$.}
\FOR{$t=0,\cdots,T$}
\STATE{sub-sample $\bm x^{(t)}$ from $\bm{x}$ with rate $r$; let $b^{(t)}$ denote the size of $\bm x^{(t)}$}
\STATE{draw $m$ samples from  $\bm z_0\sim q_0$}
\FOR{$\bm x_i\in \bm{x}^{(t)}$}
\STATE{$l_i(\lambda)= m^{-1}\sum_{j=1}^m\left\{n^{-1}\log q_0(\bm{z}_{0, j})-\log p(\bm x_{i}, F_{\lambda}(\bm{z}_{0,j}))-n^{-1}\log|\partial F_{\lambda}/\partial\bm z_{0, j}|\right\}$}
\STATE{obtain $\bm g_i^{(t)} = \nabla_{\lambda}l_i(\lambda)$ and $\bm g_i^{(t)} \leftarrow \bm g_i^{(t)}/\max(1, \|\bm g_i^{(t)}\|_2/C)$}
\ENDFOR
\STATE{obtain sanitized gradient $\bm g^{*(t)} = \left(\sum_{i=1}^{b^{(t)}}\, \bm g_i^{(t)} + \mathcal{N}(\bm 0,\sigma^{2}C^2\bm I)\right)/b^{(t)}$}
\STATE{$\lambda^{*(t+1)} = \lambda^{*(t)} - \eta\,g^{*(t)}$}
\ENDFOR
\end{algorithmic}
\end{algorithm}

\subsection{Application of DP-NF-VI to Simulated Data}\label{sec:util_DP-NF-VI}

We used a simulated dataset\footnote{We also ran the algorithm on the EHR dataset but it failed to converge.} to evaluate the utility of VI obtained via Algorithm~\ref{alg:DPNF_VI}. 
Specifically, we simulated outputs $\bm x$ as given covariates $\bm w$ and  model parameters $\bm \beta$  from the following nonlinear model
\begin{equation}\label{eqn:NL}
x_{i,k} = f(\bm{w}_i, \bm{\beta}) + r_{i} + e_{i,k},
\end{equation}
where
\begin{equation}
f(\bm{w}_i,  \bs \beta) = \beta_0 + \exp[\beta_1\, w_{i,1}] + \log[w_{i,2} + \exp(\beta_2)] + \exp[\beta_2\,w_{i,3}] + \log[w_{i,4} + \exp(\beta_4)],
\end{equation}
and, additionally, 
\begin{equation}
r_i \sim \N(0,\sigma_0^2),\quad e_{i,k} \sim \N(0, \sigma^2),
\end{equation}
for $i=1,\ldots,6000$ and $k=1,\ldots 5$. 
The covariates $\bm{w}$ were generated using $w_{i,j}\sim \mathcal{U}(0, a_j)$ independently with $i=1,\cdots, n$ and $a_1=1$, $a_2=3$, $a_3=0.5$, $a_4=2$. 
We set $\sigma_0 = \sigma = 0.2$ and $\bm{\beta} = (0.2, 1.0, 0.8, -1.2, 0.6)^T$. The random effect $r_i$ introduces correlation among 5 repeated measurements observed for the $i$-th subject, where the correlation is $ \sigma^2/( \sigma^2+ \sigma_0^2)=1/2$.

Our goal is to estimate a privacy-preserving posterior distribution of $\bm{\beta}$ and obtain its privacy-preserving posterior inference given a non-informative uniform prior $p(\bm{\beta})$ and the likelihood $l(\bm{\beta}|\bm{x})$, where, WLOG, $\sigma^2_0,\sigma^2$ are assumed known.
We applied Algorithm~\ref{alg:DPNF_VI} to the simulated data at various levels of $\mu$-GDP guarantees ($\mu=6.58, 1.12, 0.5$ and 0.27, which correspond to $\epsilon=50,5,2, 1$ when $\delta = 1\times 10^{-5}$). 
We used a MAF architecture with one MADE (1 hidden layer with 10 neurons) and one batch norm layer. 
We also used a batch of size $m=1,000$ to compute the Monte Carlo estimates of the loss function in Algorithm~\ref{alg:DPNF_VI}, a sub-sampling rate $r=500/6,000$, learning rate $\eta=0.01$ with exponential scheduler decay factor 0.999, number of iterations $T=8,000$, and clipping constant $C=10.0$.

The results on the privacy-preserving posterior inference for $\bm{\beta}$ are reported in Table~\ref{tab:bad}. 
In summary, the privacy-preserving posterior means remain roughly constant regardless of the selected privacy loss parameter $\mu$, whereas the SD values in the parentheses are greater than the corresponding non-private values (column ``original VI NF''), and become increasingly larger when $\mu$ decreases.
Additionally, the private correlations among different elements of $\bm\beta$ computed using DP-NF-VI are in general more pronounced with respect to the corresponding original values and are associated with large SDs from the 5 different DP-NF-VI runs, indicating a general lack of stability for these estimates.
\begin{table}[!htb]
\centering
\caption{Differentially private VI on the nonlinear regression model parameters $\bm{\beta}$ from DP-NF-VI (Algorithm~\ref{alg:DPNF_VI}).}\label{tab:bad}
\resizebox{1.0\textwidth}{!}{%
\begin{tabular}{l|c@{\hspace{3pt}}c|cccc}
  \toprule
 {parameter} & \makecell[c]{{original}\\{nonlinear}\\{model}}& \makecell[c]{{original }\\{NF-VI }} & \makecell[c]{{${\mu=6.68}$}\\{${\sigma=1.30}$}} & \makecell[c]{{${\mu=1.12}$}\\{${\sigma=6.68}$}} & \makecell[c]{{${\mu=0.50}$}\\{${\sigma=14.88}$}} & \makecell[c]{{${\mu=0.27}$}\\{${\sigma=27.82}$}}\\
\midrule
\multicolumn{6}{c}{Posterior Mean}\\
\midrule
$\beta_0$\; ( 0.2)	&\ 0.1581	&\ 0.1593 (\ 0.0020)	&\ 0.1532 (\ 0.0035)	&\ 0.1381 (\ 0.0151)	&\ 0.1231 (\ 0.0486)	&\ 0.1419 (\ 0.0701)	\\
$\beta_1$\; ( 1.0)	&\ 1.0022	&\ 1.0016 (\ 0.0015)	&\ 0.9952 (\ 0.0040)	&\ 0.9957 (\ 0.0036)	&\ 0.9922 (\ 0.0059)	&\ 0.9943 (\ 0.0026)	\\
$\beta_2$\; ( 0.8)	&\ 0.8262	&\ 0.8265 (\ 0.0013)	&\ 0.8386 (\ 0.0034)	&\ 0.8404 (\ 0.0150)	&\ 0.8378 (\ 0.0208)	&\ 0.8584 (\ 0.0278)	\\
$\beta_3$\; (-1.2)	&-1.1909	&-1.1920 (\ 0.0008)	&-1.1928 (\ 0.0019)	&-1.1891 (\ 0.0079)	&-1.1822 (\ 0.0201)	&-1.1945 (\ 0.0246)	\\
$\beta_4$\; ( 0.6)	&\ 0.6088	&\ 0.6092 (\ 0.0006)	&\ 0.6100 (\ 0.0010)	&\ 0.6215 (\ 0.0057)	&\ 0.6311 (\ 0.0390)	&\ 0.6057 (\ 0.0479)	\\
\midrule
\multicolumn{6}{c}{Posterior SD}\\
\midrule
$\beta_0$	&\ 0.0265	&\ 0.0255 (\ 0.0011)	&\ 0.1341 (\ 0.0147)	&\ 0.2576 (\ 0.0241)	&\ 0.2573 (\ 0.0490)	&\ 0.2791 (\ 0.0825)	\\
$\beta_1$	&\ 0.0036	&\ 0.0036 (\ 0.0001)	&\ 0.1089 (\ 0.0884)	&\ 0.0687 (\ 0.0587)	&\ 0.1344 (\ 0.1247)	&\ 0.0554 (\ 0.0058)	\\
$\beta_2$	&\ 0.0179	&\ 0.0174 (\ 0.0008)	&\ 0.0755 (\ 0.0222)	&\ 0.1267 (\ 0.0250)	&\ 0.1729 (\ 0.0309)	&\ 0.1655 (\ 0.0305)	\\
$\beta_3$	&\ 0.0151	&\ 0.0151 (\ 0.0004)	&\ 0.0812 (\ 0.0056)	&\ 0.1341 (\ 0.0141)	&\ 0.1644 (\ 0.0203)	&\ 0.1581 (\ 0.0217)	\\
$\beta_4$	&\ 0.0192	&\ 0.0191 (\ 0.0006)	&\ 0.0889 (\ 0.0163)	&\ 0.1197 (\ 0.0089)	&\ 0.1595 (\ 0.0384)	&\ 0.2012 (\ 0.0399)	\\
\midrule
\multicolumn{6}{c}{Posterior Correlation}\\
\midrule
$\beta_0, \beta_1$	&-0.1200	&-0.0319 (\ 0.0499)	&-0.1147 (\ 0.2240)&-0.1230 (\ 0.2623)	&\ 0.0237 (\ 0.1028)	&-0.1619 (\ 0.2592)	\\
$\beta_0, \beta_2$	&-0.4340	&-0.3415 (\ 0.1899)	&-0.1180 (\ 0.5120)&-0.3923 (\ 0.4756)	&\ 0.0396 (\ 0.3675)	&-0.3269 (\ 0.7675)	\\
$\beta_0, \beta_3$	&-0.7370	&-0.7550 (\ 0.0127)	&-0.9108 (\ 0.0851)&-0.8923 (\ 0.0353)	&-0.8629 (\ 0.0407)	&-0.8622 (\ 0.0420)	\\
$\beta_0, \beta_4$	&-0.4880	&-0.4883 (\ 0.0356)	&-0.5299 (\ 0.4673)&-0.8545 (\ 0.0839)	&-0.6826 (\ 0.1586)	&-0.4920 (\ 0.5249)	\\
$\beta_1, \beta_2$	&-0.0030	&-0.0082 (\ 0.0417)	&-0.0310 (\ 0.5503)&\ 0.0285 (\ 0.0406)	&\ 0.0916 (\ 0.1965)	&-0.0395 (\ 0.1101)	\\
$\beta_1, \beta_3$	&-0.0160	&-0.0892 (\ 0.0264)	&\ 0.1255 (\ 0.2115)&-0.0381 (\ 0.2270)	&-0.1815 (\ 0.1485)	&-0.0691 (\ 0.2162)	\\
$\beta_1, \beta_4$	&-0.0080	&-0.0680 (\ 0.0252)	&-0.2880 (\ 0.4993)&\ 0.1445 (\ 0.2869)	&-0.0457 (\ 0.0595)	&\ 0.0784 (\ 0.2416)	\\
$\beta_2, \beta_3$	&\ 0.0040	&-0.0538 (\ 0.1526)	&-0.0947 (\ 0.6286)&\ 0.1131 (\ 0.4442)	&-0.2520 (\ 0.4011)	&\ 0.1453 (\ 0.7946)	\\
$\beta_2, \beta_4$	&-0.0020	&-0.0836 (\ 0.0936)	&-0.2267 (\ 0.4445)&\ 0.3336 (\ 0.5772)	&-0.2036 (\ 0.2303)	&-0.0271 (\ 0.5412)	\\
$\beta_3, \beta_4$	&-0.0020	&\ 0.0318 (\ 0.0443)	&\ 0.3368 (\ 0.3616)&\ 0.6469 (\ 0.1602)	&\ 0.4218 (\ 0.0884)	&\ 0.2300 (\ 0.5860)	\\
\midrule
\multicolumn{7}{l}{The results were averaged over 5 runs of DP-NF-VI; the SD values over the 5 runs in the parentheses.}\\
\bottomrule
\end{tabular}\vspace{-9pt}
}
\end{table}

As a comparison, we also performed VI for $\bm \beta$ in the nonlinear model based on privacy-preserving  synthetic  data generated using DP-NF for density estimation (Algorithm~\ref{alg:DPNF_density}). 
For this task, we use a MAF architecture consisting of one MADE (1 hidden layer, 10 neurons) and one batch norm layer. 
We also used RMSprop with a batch size of 500, a learning rate of 0.01, and an exponential learning rate scheduler with a decay factor equal to 0.999. The total number of iterations was set to 8,000 and the gradient clipping constant $C$ to 10.0. 

The results are presented in Table~\ref{tab:good}. In general, regardless of $\mu$, the posterior estimates of $\bm \beta$  from NF-VI based on privacy-preserving synthetic data show superior accuracy and are more similar to the results from non-private inference than those obtained directly from DP-NF-VI (Table~\ref{tab:bad}). 
In addition, the SD values in the parentheses across multiple sets of synthetic data are relatively small, while the  posterior correlations $\bm \beta$  are reasonably close to their original non-private values.

\begin{table}[!htb]
\centering
\caption{Privacy-preserving VI based on synthetic data generated via DP-NF for density estimation (Algorithm~\ref{alg:DPNF_density}).}\label{tab:good}
\resizebox{1.0\textwidth}{!}{%
\begin{tabular}{l|cc|cccc}
  \toprule
 {parameter} & \makecell[c]{{original}\\{ nonlinear}\\{model}}& \makecell[c]{{original }\\{ NF }} & \makecell[c]{{${\mu=6.68}$}\\{${\sigma=1.30}$}} & \makecell[c]{{${\mu=1.12}$}\\{${\sigma=6.68}$}} & \makecell[c]{{${\mu=0.50}$}\\{${\sigma=14.88}$}} & \makecell[c]{{${\mu=0.27}$}\\{${\sigma=27.82}$}}\\
\midrule
\multicolumn{6}{c}{Posterior Mean}\\
\midrule
$\beta_0$ ( 0.2)	&\ 0.1581	&\ 0.1593 (\ 0.0020)	&\ 0.1632 (\ 0.0146)	&\ 0.1062 (\ 0.0505)	&\ 0.0866 (\ 0.0127)	&\ 0.0448 (\ 0.0600)	\\
$\beta_1$ ( 1.0)	&\ 1.0022	&\ 1.0016 (\ 0.0015)	&\ 0.9891 (\ 0.0060)	&\ 0.9943 (\ 0.0060)	&\ 0.9917 (\ 0.0097)	&\ 0.9775 (\ 0.0126)	\\
$\beta_2$ ( 0.8)	&\ 0.8262	&\ 0.8265 (\ 0.0013)	&\ 0.8235 (\ 0.0078)	&\ 0.8357 (\ 0.0286)	&\ 0.8246 (\ 0.0230)	&\ 0.8294 (\ 0.0515)	\\
$\beta_3$ (-1.2)	&-1.1909	&-1.1920 (\ 0.0008)	&-1.1807 (\ 0.0081)	&-1.1630 (\ 0.0251)	&-1.1406 (\ 0.0138)	&-1.1034 (\ 0.0225)	\\
$\beta_4$ ( 0.6)	&\ 0.6088	&\ 0.6092 (\ 0.0006)	&\ 0.6011 (\ 0.0230)	&\ 0.6261 (\ 0.0237)	&\ 0.6197 (\ 0.0145)	&\ 0.6214 (\ 0.0233)	\\
\midrule
\multicolumn{6}{c}{Posterior SD}\\
\midrule
$\beta_0$	&\ 0.0265	&\ 0.0255 (\ 0.0011)	&\ 0.0280 (\ 0.0004)	&\ 0.0279 (\ 0.0003)	&\ 0.0277 (\ 0.0006)	&\ 0.0274 (\ 0.0004)	\\
$\beta_1$	&\ 0.0036	&\ 0.0036 (\ 0.0001)	&\ 0.0040 (\ 0.0000)	&\ 0.0040 (\ 0.0000)	&\ 0.0039 (\ 0.0001)	&\ 0.0038 (\ 0.0001)	\\
$\beta_2$	&\ 0.0179	&\ 0.0174 (\ 0.0008)	&\ 0.0201 (\ 0.0003)	&\ 0.0210 (\ 0.0005)	&\ 0.0207 (\ 0.0013)	&\ 0.0203 (\ 0.0017)	\\
$\beta_3$	&\ 0.0151	&\ 0.0151 (\ 0.0004)	&\ 0.0159 (\ 0.0003)	&\ 0.0153 (\ 0.0002)	&\ 0.0144 (\ 0.0007)	&\ 0.0142 (\ 0.0006)	\\
$\beta_4$	&\ 0.0192	&\ 0.0191 (\ 0.0006)	&\ 0.0208 (\ 0.0002)	&\ 0.0229 (\ 0.0003)	&\ 0.0247 (\ 0.0020)	&\ 0.0250 (\ 0.0020)	\\
\midrule
\multicolumn{6}{c}{Posterior Correlation}\\
\midrule
$\beta_0, \beta_1$	&-0.1200	&-0.0319 (\ 0.0499)	&-0.0590 (\ 0.0262)&-0.0568 (\ 0.0130)	&-0.0647 (\ 0.0338)	&-0.0359 (\ 0.0185)	\\
$\beta_0, \beta_2$	&-0.4340	&-0.3415 (\ 0.1899)	&-0.4184 (\ 0.0159)&-0.4155 (\ 0.0378)	&-0.4588 (\ 0.0606)	&-0.4774 (\ 0.0296)	\\
$\beta_0, \beta_3$	&-0.7370	&-0.7550 (\ 0.0127)	&-0.7293 (\ 0.0050)&-0.6919 (\ 0.0092)	&-0.6951 (\ 0.0200)	&-0.7049 (\ 0.0326)	\\
$\beta_0, \beta_4$	&-0.4880	&-0.4883 (\ 0.0356)	&-0.5206 (\ 0.0103)&-0.5418 (\ 0.0122)	&-0.5168 (\ 0.0315)	&-0.4835 (\ 0.0388)	\\
$\beta_1, \beta_2$	&-0.0030	&-0.0082 (\ 0.0417)	&-0.0289 (\ 0.0211)&-0.0244 (\ 0.0196)	&-0.0315 (\ 0.0297)	&-0.0774 (\ 0.0207)	\\
$\beta_1, \beta_3$	&-0.0160	&-0.0892 (\ 0.0264)	&-0.0723 (\ 0.0285)&-0.0674 (\ 0.0176)	&-0.0607 (\ 0.0249)	&-0.0557 (\ 0.0267)	\\
$\beta_1, \beta_4$	&-0.0080	&-0.0680 (\ 0.0252)	&-0.0269 (\ 0.0204)&-0.0318 (\ 0.0166)	&-0.0229 (\ 0.0198)	&-0.0428 (\ 0.0171)	\\
$\beta_2, \beta_3$	&\ 0.0040	&-0.0538 (\ 0.1526)	&-0.0202 (\ 0.0194)&-0.0409 (\ 0.0257)	&\ 0.1278 (\ 0.2343)	&\ 0.2144 (\ 0.2168)	\\
$\beta_2, \beta_4$	&-0.0020	&-0.0836 (\ 0.0936)	&-0.0346 (\ 0.0156)&-0.0518 (\ 0.0347)	&-0.1289 (\ 0.0761)	&-0.1792 (\ 0.0979)	\\
$\beta_3, \beta_4$	&-0.0020	&\ 0.0318 (\ 0.0443)	&\ 0.0484 (\ 0.0110)&\ 0.0316 (\ 0.0170)	&-0.0257 (\ 0.0677)	&-0.0676 (\ 0.0510)	\\
\hline
\multicolumn{7}{l}{The results were averaged over 5 sets of synthetic data generated via DP-NF for density estimation; the SD values}\\
\multicolumn{7}{l}{in the parentheses are computed over the 5 sets.}\\
\bottomrule
\end{tabular}%
}
\end{table}

\subsection{Further Utility Analysis of DP-NF-VI}

The unsatisfactory results on the posterior correlations among the parameters, as displayed in Table \ref{tab:bad}, are not accidental. We tested the DP-NF-VI algorithm in multiple experiments on different datasets, and similar results were observed. We conjecture that this might be due to the usage of the free energy bound as the loss function in DP-NF-VI. 
Minimizing the free energy bound in NF-VI is equivalent to minimizing the KL divergence between the true posterior and its variational approximation. 
As a result of noise injection and gradient clipping in differentially private gradient-based optimization to achieve DP guarantees, the posterior variances of the parameters are expected to increase in the  privacy-preserving variational distribution relative to those in the non-private target distribution. 
Thus, the  DP-NF-VI procedure looks for densities in the variational family that are the most similar to the true posterior measured by the KL divergence but have larger variances. 
This may lead to an optimal variational density with a different correlation structure than the original. 

\begin{figure}[!htb]
\centering
\includegraphics[trim= 2cm 1.0cm 2cm 0.8cm,clip,width=\linewidth]{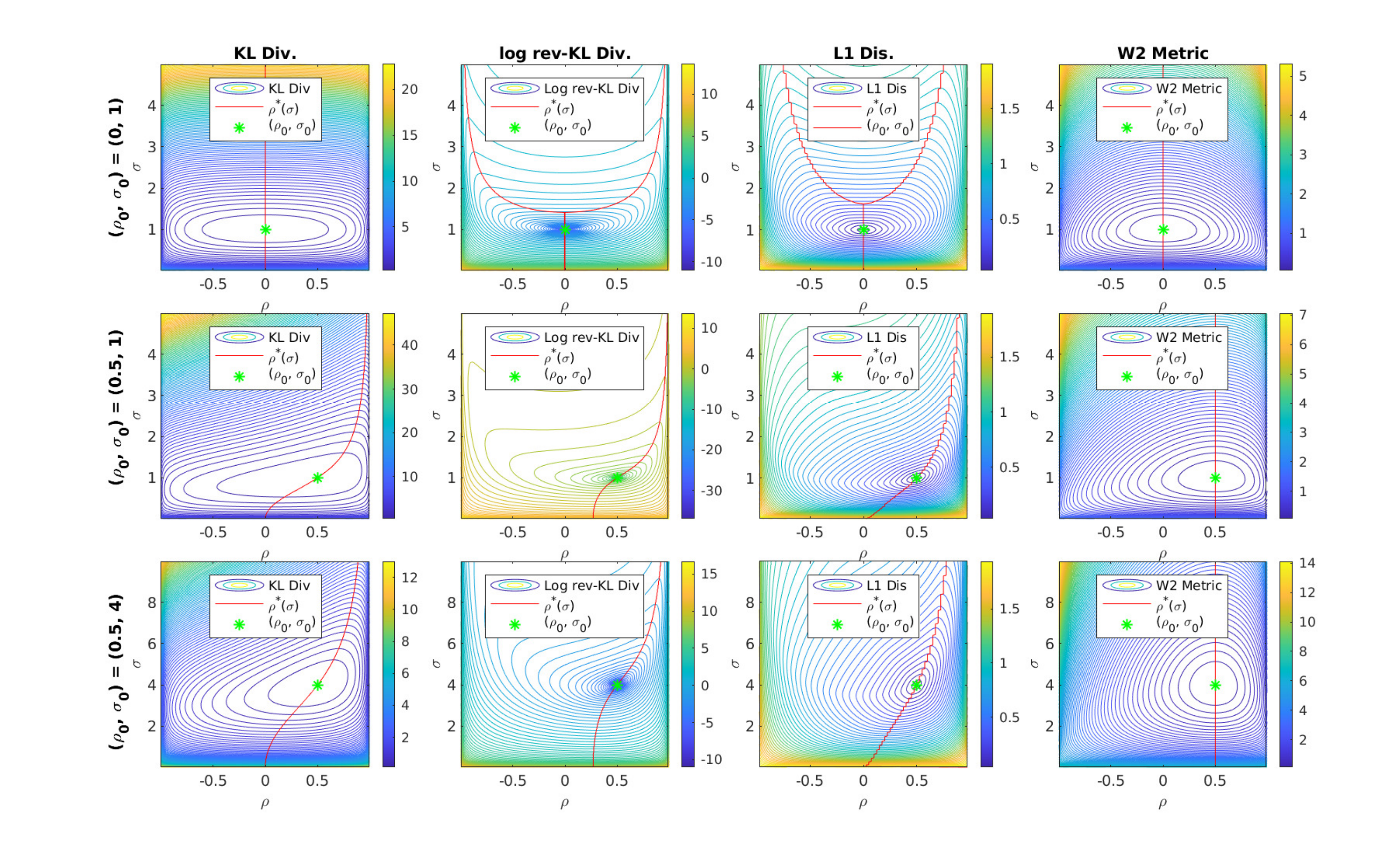}
\caption{KL Divergence, reversed KL divergence, $\ell_1$ and Wasserstain distance between \emph{target} distribution $p=\mathcal{N}(\bm 0, \Sigma_0)$ and its \emph{private} approximation $q=\mathcal{N}(\bm 0, \Sigma)$. The red line in contour plots is the trajectory of $\rho^*(\widehat{\sigma})={\arg\min}_{\rho}\,L(\rho_0, \sigma_0, \rho, \widehat{\sigma})$ that minimizes loss function $L$ given $\widehat{\sigma}$.}\label{fig:Diagram}
\end{figure}

To further demonstrate this phenomenon and test our conjecture, we run the following experiment. Despite the simplicity of the experiment, the results  are interesting and insightful.
Suppose the \emph{target} distribution is a bivariate Gaussian distribution 
\[
p = \mathcal{N}(\bm 0, \bm{\Sigma}_0),\,\,\text{with}\,\,\bm{\Sigma}_0 = \sigma_0^2\,\left[(1-\rho_0)\bm{I}_{2\times 2} + \rho_0 \bm{1}_{2\times 2}\right].
\]
We aim to approximate the distribution by another bivariate Gaussian distribution
\[
q = \mathcal{N}(\bm 0, \Sigma),\,\,\text{with}\,\,\bm{\Sigma} = \sigma^2\,\left[(1-\rho)\,\bm{I}_{2\times 2} + \rho \bm{1}_{2\times 2}\right],
\]
which can be regarded as the \emph{private} distribution, expected to have a larger marginal variance in each dimension compared to those in $p$; i.e., $\sigma > \sigma_{0}$. 
We examine the following four dissimilarity (or divergence) metrics between $p$ and $q$
\[
L(p,q)=
\begin{cases}
\mbox{KL divergence} & D(q\|p),\\
\mbox{reversed KL divergence} & D(p\|q),\\
\mbox{$\ell_1$ distance\footnotemark} & \|q- p\|_1 = \int_{\mathbb{R}^2}|p(\bm x)-q(\bm x)|\,d\bm x,\\
\mbox{Wasserstein distance} & W_{\alpha}(p, q) = (\inf_{\gamma \in \Gamma(p, q)}\mathbb{E}_{\gamma}|p-q|^\alpha)^{1/\alpha}.
\end{cases}.
\]
\addtocounter{footnote}{-1}
\footnotetext{Equivalent to twice the total variation norm $\|q-p\|_{TV} = \sup_{A\subset \mathbb{R}^2}\,|p(A) - q(A)|$}
In this experiment, the Wasserstein distance is computed as
\[
W_2(p, q) = \text{trace}(\bm{\Sigma}_0 + \bm{\Sigma} - 2[\bm{\Sigma}^{1/2}\,\bm{\Sigma}_0\,\bm{\Sigma}^{1/2}]^{1/2}).
\]
We computed the dissimilarity between $p$ and $q$ (loss function) for each of the 4 metrics above as a function of $\rho$ and $\sigma$ and generated the corresponding contour plots in Figure~\ref{fig:Diagram}, for three pairs $(\rho_0, \sigma_0)$ at $(0, 1), (1/2, 1)$ and $(1/2, 4)$, respectively. 
We then identified the value of $\rho$ that minimizes the dissimilarity given an inflated standard deviation $\sigma=\widehat{\sigma}$; that is, 
\[
\rho^*(\widehat{\sigma})={\arg\min}_{\rho} L(\rho_0, \sigma_0, \rho, \widehat{\sigma}),
\]


The results of the experiment in Figure~\ref{fig:Diagram} suggest that the correlation $\rho$ of the private distribution with \emph{inflated} variance ($\sigma > \sigma_0$) that optimally approximates the target distribution (in the sense of minimal dissimilarity) depends, in general, on both $\rho_0$ and the selected  dissimilarity  metric.

When $\rho_0=0$, the optimal private approximation determined through both the KL divergence and the Wasserstein distance correctly matches the target correlation, i.e., $\rho=0$;
if the reverse KL divergence or the $\ell_1$ distance is used, the optimal private approximation is not even unique, until its marginal variance $\sigma$ is sufficiently close or smaller than $\sigma_{0}$.  

When $\rho_0$ is not 0, the optimal private approximation with inflated variance is characterized by a correlation $\rho$ that rapidly tends to 1.0 as the inflation ratio $\sigma/\sigma_{0}$ increases. This behavior is observed in the cases of the KL divergence, the reverse KL divergence, and the $\ell_1$ distance. 
However, the Wasserstein distance is still able to produce a private approximation with $\rho=\rho_0$ independently of the inflated variance $\sigma$. 

In conclusion, the Wasserstein distance appears to be the best metric to use if one aims to preserve the correlation structure among the estimated parameters in private variational distributions, even with inflated marginal variance resulting from noise injection to achieve privacy guarantees.
%

\section{Discussion}\label{sec:discussion}

We examined the feasibility and utility of generating synthetic data via privacy-preserving NF for density estimation, with an application to an EHR dataset. 
Overall, the results suggest that  the downstream learning  (prediction) and inferential procedure (VI) based on the privacy-preserving synthetic data can yield results of satisfactory and acceptable utility at practically reasonable choices of privacy loss parameters for the dataset. 
Though more empirical studies will be needed, this study on a real dataset with a relatively small sample size and a relatively large number of attributes of mixed types and missing values, suggests generating  synthetic data via privacy-preserving NF for density estimation may be a practically feasible option for releasing and sharing sensitive information.

We also examined the feasibility of directly sanitizing the NF-VI procedure to generate privacy-preserving posterior inference. 
We show that the free energy bound as the loss function for NF-VI may yield variational distribution deviating significantly from the target distribution. 
This implies that alternative loss functions may better preserve the information on certain aspects of the target distribution, e.g., its correlation structure.
In this context, we plan to further explore how the use of different distance metrics affects the performance of algorithms for DP-NF-VI, as a way to develop more effective strategies leveraging, for example, Wasserstein distance-based loss formulations.

The code for both the density estimation and VI through DP-NF is available on Github at \url{https://github.com/cedricwangyu/DPNF}. 

\small{
\section*{Acknowledgements}
This work is supported by an NSF Big Data Science \& Engineering grant \#1918692 (PI DES), an NSF CAREER grant \#1942662 (PI DES), an NSF OAC CDS\&E grant \#2104831 (Notre Dame PI DES) and used computational resources provided through the Center for Research Computing at the University of Notre Dame.
}

\section*{Appendix}
\textbf{A. Compliance With Ethical Standards}

The EHR dataset utilized in this work contains data from external studies that involved human participants.  The medical procedures that occurred in these studies were performed in accordance with both the ethical standards of the institutional and national research committee and with the 1964 Helsinki declaration and its later amendments. This study is classified as research not involving human subjects and was approved on June 13th, 2019, by the Office of Research Compliance and Institutional Review Board at the University of Notre Dame under IRB\#19-05-5371.

\textbf{B. Acronyms in  CVSim-6 model}
\begin{table}[!htb]
\centering
\caption{Acronyms in  CVSim-6 model}\label{tab:acro}
\resizebox{0.9\textwidth}{!}{  
\begin{tabular}{llll}
\toprule
\multicolumn{4}{c}{\bf Flows}\\
\midrule
$q_{li}$ & Left ventricular inflow & $q_{lo}$ & Left ventricular outflow\\
$q_{ri}$ & Right ventricular inflow & $q_{a}$ & Systemic aortic flow\\
$q_{ro}$ & Right ventricular outflow & $q_{pv}$ & Pulmonary venous flow\\
\midrule
\multicolumn{4}{c}{\bf Pressures}\\
\midrule
$P_{pv}$ & Pulmonary venous pressure & $P_{l}$ & Left ventricular pressure\\
$P_{a}$ & Systemic arterial pressure & $P_{v}$ & Systemic venous pressure\\
$P_{r}$ & Right ventricular pressure & $P_{pa}$ & Pulmonary arterial pressure\\
$P_{th}$ & Thransthoracic pressure & $P_{r,dias}$ & Right ventricular diastolic pressure\\
$P_{r,sys}$ & Right ventricular systolic pressure & $P_{l,dias}$ & Left ventricular diastolic pressure\\
$P_{l,sys}$ & Left ventricular systolic pressure\\
\midrule
\multicolumn{4}{c}{\bf Resistances}\\
\midrule
$R_{a}$ & Systemic arterial resistance & $R_{v}=R_{ri}$ & Systemic venous resistance\\
$R_{lo}$ & Aortic valve resistance & $R_{v}$ & Systemic venous resistance\\
$R_{ro}$ & Pulmonary valve resistance & $R_{pv}=R_{li}$ & Pulmonary venous resistance\\
\midrule
\multicolumn{4}{c}{\bf Capacitances}\\
\midrule
$C_{l}(t)$ & Left ventricular capacitance & $C_{r}(t)$ & Right ventricular capacitance\\
$C_{a}$ & Systemic arterial capacitance & $C_{v}$ & Systemic venous capacitance\\
$C_{pa}$ & Pulmonary arterial capacitance & $C_{pv}$ & Pulmonary venous capacitance\\
$C_{l,dias}$ & Diastolic left ventricular capacitance & $C_{l,sys}$ & Systolic left ventricular capacitance\\
$C_{r,dias}$ & Diastolic right ventricular capacitance & $C_{r,sys}$ & Systolic \\
\midrule
\multicolumn{4}{c}{\bf Times}\\
\midrule
$T_{tot}$ & Total heart cycle time & $T_{dias}$ & Duration of diastolic phase\\
$T_{sys}$ & Duration of systolic phase\\
\midrule
\multicolumn{4}{c}{\bf Volumes}\\
\midrule
$V_{tot}$ & Total blood volume & $V_{0,tot}$ & Unstressed blood volume\\
\bottomrule
\end{tabular}}
\end{table}

\textbf{C. Hyper-parameters}

We list the hyper-parameters used in all of our experiments in Table~\ref{tab:hyper}. 
EHR DP-DE indicates the differentially private density estimation in the EHR data (see Sec~\ref{sec:private_surrogate}); EHR VI means VI with KL divergence in the privacy-preserving synthetic EHR data (see Sec~\ref{sec:DP-VI-syn-EHR}); Reg DP-VI represents differentially private VI in the simulated data via the nonlinear regression model (see Sec~\ref{sec:util_DP-NF-VI}); Reg DP-DE means differentially private density estimation of the simulated data (see Sec~\ref{sec:util_DP-NF-VI}); Reg VI means VI with KL divergence in the privacy-preserving simulated data l (See Sec~\ref{sec:util_DP-NF-VI}). Except for the surrogate model (a fully connected neural network) for the CVSim-6 model, where the activation function tanh was used, ReLU was the activation function in all other NF and neural networks.
\begin{table}[!htb]
\centering
\caption{Hyper-parameters used in all the experiments}\label{tab:hyper}
\resizebox{0.8\textwidth}{!}{  
\begin{tabular}{c|ccccc}
\toprule
parameter & EHR DP-DE & EHR VI & Reg DP-VI & Reg DP-DE & Reg VI\\
\midrule
block No. & 15 & 5 & 1 & 18 & 1\\
hidden No. & 1 & 1 & 1 & 1 & 1\\
hidden size & 200 & 100 & 10 & 100 & 10\\
input size & 19 & 2 & 5 & 9 & 5\\
iter No. & 8000 & 15000 & 8000 & 8000 & 8000\\
batch size & 100 & 500 & 1000 & 100 & 1000\\
optimizer & SGD & RMSProp & RMSProp & RMSProp & RMSProp\\
learn rate & 0.00002 & 0.01 & 0.01 & 0.002 & 0.01\\
scheduler & - & Exp & Exp & Exp & Exp\\
decay factor & - & 0.9999 & 0.999 & 0.9995 & 0.999\\
clipping & 10 & - & 10 & 5 & -\\
poisson rate & 50\% & - & 8.33\% & 8.33\% & -\\
\bottomrule
\end{tabular}}
\end{table}

\bibliographystyle{abbrvnat}
\bibliography{ref}

\end{document}